\documentclass[journal,12pt,onecolumn]{IEEE_Transaction}
\usepackage[top=2.0cm, bottom=2.2cm, left=2.2cm,right=2.2cm]{geometry}%Makes 2.5cm margins on my printer
\ifCLASSOPTIONonecolumn
\usepackage{setspace}
\fi
\usepackage{amsmath}
\usepackage{amssymb}
\usepackage{amsthm}
\usepackage{eqnarray}
\usepackage{booktabs}
\usepackage{graphicx}
\usepackage{epsfig,color}
\usepackage{wrapfig}
\usepackage{algorithmic}
\usepackage{cite}
\usepackage{comment}
\usepackage{subcaption}
\usepackage[font=small,labelfont=bf]{caption}
\usepackage{xcolor}%\textcolor{blue}{}. Contents of {} are limited to no more than a paragraph. Many color names work.

\usepackage{etoolbox}
\makeatletter
\patchcmd{\@makecaption}
  {\scshape}
  {}
  {}
  {}
\makeatother
\def\tablename{Table}
\renewcommand{\thetable}{\arabic{table}} %Changes Table numbering from roman to arabic

\newcommand{\subparagraph}{}
\usepackage[compact]{titlesec}
\titlespacing{\section}{0pt}{0.9ex}{0.7ex}
\titlespacing{\subsection}{0pt}{0.7ex}{0.7ex}
\titlespacing{\subsubsection}{0pt}{0.7ex}{0.7ex}

\newcommand{\mycomment}[1]{%
}%
\usepackage{environ}
\NewEnviron{nestedcomment}{\mycomment{\BODY}}
\def\urltilde{\kern -.15em\lower .7ex\hbox{\~{}}\kern .04em}

\allowdisplaybreaks

\begin{document}
\bstctlcite{IEEEexample:BSTcontrol}

\title{Machine Learning in NextG Networks via \\ Generative Adversarial Networks}

\author{Ender Ayanoglu,~\IEEEmembership{Fellow,~IEEE}, Kemal Davaslioglu,~\IEEEmembership{Member, IEEE},\\ Yalin E. Sagduyu,~\IEEEmembership{Senior Member, IEEE}% <-this % stops a space
\thanks{E.~Ayanoglu is with the Center for Pervasive Communications and Computing (CPCC), Dept. of EECS, UC Irvine, CA, USA, K.~Davaslioglu is with UTS Inc., MD, USA, and Y.~E.~Sagduyu is with Intelligent Automation, A BlueHalo Company, MD, USA.
}
% <-this % stops a space
}

\maketitle
\vspace{-0.5in}
\begin{abstract}
Generative Adversarial Networks (GANs) are Machine Learning (ML)
algorithms that have the ability to address competitive resource
allocation problems together with detection and mitigation of
anomalous behavior. In this paper, we investigate their use in
next-generation (NextG) communications within the context of cognitive
networks to address {\em i)\/} spectrum sharing, {\em ii)\/} detecting anomalies, and
{\em iii)\/} mitigating security attacks.
GANs
have the following advantages. First, they can learn and synthesize
field data, which can be costly, time consuming, and nonrepeatable. Second, they enable pre-training classifiers by using
semi-supervised data. Third, they facilitate increased
resolution. Fourth, they enable the recovery of corrupted bits in the
spectrum.
The paper provides the basics of GANs, a comparative discussion on different kinds of GANs, performance measures for GANs in computer vision and image processing as well as wireless applications, a number of datasets for wireless applications, performance measures for general classifiers, a survey of the literature on GANs for {\em i)--iii)\/} above,
and future research directions.
As a use case of GAN for nextG communications, we show that a GAN can be effectively applied for anomaly detection in signal classification (e.g., user authentication) outperforming another state-of-the-art ML technique such as an autoencoder.
\end{abstract}

\begin{IEEEkeywords}
Generative adversarial networks (GANs), conditional GANs, generative modeling, spectrum sharing, anomaly
detection, outlier detection, wireless security, unsupervised learning.
\end{IEEEkeywords}

% Will be ignored for other modes.
\IEEEpeerreviewmaketitle
\section{Introduction}\label{sec:introduction}
The number of radios and their use are increasing exponentially. Over the last several years, wireless data transmission has grown up by approximately 50\% per year \cite{Tilghman19-2,Tilghman19}. This increase in demand has largely been driven by novel applications such as streaming videos and social media on smart devices, and has strained the ability of the limited available wireless spectrum to support it. The coming era of the Internet-of-Things (IoT) and its anticipated goal of connecting tens of billions of devices via wireless will make this situation even more challenging for next-generation wireless communication networks (NextG). Classical methods for medium access control and physical layer operation, designed to handle a relatively small number of users in a given space-time-frequency resource, will soon be overwhelmed and alternatives must be sought.

Wireless systems of the future are anticipated to share the available spectrum rather than operating with exclusive assigned frequencies. It is generally expected that this sharing will be across all possible dimensions, including space, time, and frequency, and will involve a huge quantity of interactions among a very large number of radios. Flexible methods will be needed to efficiently use the limited available resources, quickly adapting themselves to changing environments and Quality-of-Service (QoS) requirements. This will be true even as wireless systems move to higher frequencies (such as anticipated for mmWave and THz bands) where available bandwidth is more abundant. Due to shorter propagation distances and the increased prevalence of blockages, it will require networks to continually reconfigure themselves via handoffs, cooperative relaying, beamforming, and so on. Given the exponentially increasing demand for wireless connectivity and the inevitable increase in the complexity of the networks that will supply it, it is no surprise that research is turning towards data-driven techniques of Artificial Intelligence (AI) and Machine Learning (ML) techniques for help in addressing these issues by learning from the spectrum data and solve complex tasks.

The Defense Advanced Research Projects Agency (DARPA) recently built a testbed named {\em Colosseum\/} for testing shared-spectrum communication systems \cite{Tilghman19-2,Tilghman19}. The vision is to emulate more than 65,000 unique interactions, such as text messages or video streams, among 128 radios at once. This testbed is built with the full expectation that the type of complicated spectrum sharing scenario described above is likely and can be efficiently managed by AI \cite{Tilghman19-2,Tilghman19}.

There are two possible ways of sharing spectrum. One is an open access model, that is similar to an unlicensed band, such as the Industrial, Scientific, and Medical (ISM) band. A more desirable model from an interference viewpoint is known as the hierarchical access model, or the cognitive radio network. When the cognitive radio concept was first proposed, learning was already considered to be essential for its operation \cite{MM99,Mitola00}. There are three reasons why considering ML for cognitive radio is important. First, as discussed in \cite{Tilghman19-2,Tilghman19}, due to the expected numbers of wireless devices and the complexity of the services they provide, the learning process suggested in \cite{MM99,Mitola00} cannot be managed by classical model-based techniques, and will need to employ ML algorithms. Second, for complex radios with many inputs and many outputs, a very large number of actions will be necessary to account for all possible radio states. In the past, most cognitive radio research focused on radios that are hard-coded to manage these states. ML provides an opportunity for a learning engine to auto-generate these actions, rather than pre-programming them \cite{CHSO07}. Third, ML provides a framework in which to incorporate memory from past actions and results into current operations, and thus more quickly adapt to changing conditions in the future \cite{CHSO07}.

As ML techniques rely on the availability of representative data, it has become imperative to generate and curate data samples for training and testing of spectrum operations. To that end, synthetic data can be generated to help with ML training and testing processes, when there is a lack of sufficient number of real data samples. Building upon recent advances in deep neural networks, Generative Adversarial Networks (GANs) have been originally introduced to generate synthetic data that cannot be distinguished from real data \cite{GPMXWOCB14}. One direct use case of GANs is to generate synthetic training data samples for data augmentation purposes to strengthen the training process of ML algorithms. Later, the use of GANs has been extended to support domain adaptation, defend against adversarial attacks, enable unsupervised learning of data, detect anomalies embedded in rich data representations, and security applications. As far as defending against adversarial attacks, GANs are capable of generating adversarial samples. These can be used for network attacks that may be missed by the ML detector. Or, they can be added to the training set to help the ML detector adjust its classification boundary in order to acquire better detection capabilities.

While these use cases also apply in wireless communication systems, their effective application hinges upon careful account of wireless communication characteristics:
\begin{enumerate}
    \item  Training data for wireless communications is typically limited. There are sensing limitations regarding sampling rate (due to hardware effects) and time spent for sensing (balancing the sensing-communication tradeoff). Therefore, GANs can be used to augment the training data for wireless applications, such as spectrum sensing and wireless signal detection and classification (e.g., jammer identification) when the Radio Frequency (RF) data typically involves uncertainties due to noise, channel, traffic, and interference effects \cite{ESS18}.
    \item Wireless data is heavily environment dependent. For example, the training data collected in a laboratory environment (e.g., indoor) does not necessarily match channel characteristics in test time (e.g., outdoor). Therefore, the adaptation of test or training data to channel conditions raises the need for domain adaptation with GANs \cite{DS18}.
    \item Signal spoofing is a key element of wireless security that poses threats for infiltration through signal/user authentication systems, and can be alternatively used to set up decoys/honeypots from the defense point of view \cite{SDS19, SDS21}.
    \item Wireless communication systems are subject to attacks due to the open nature of wireless spectrum and anomalies due to hardware impairments and intended/unintended interference. GANs can be used to protect wireless communications against attacks and detect anomalies. %(see Section~\ref{sec:use-case} for results on anomaly detection).
\end{enumerate}

The unique characteristics of wireless communication systems that need to be considered for the GAN applications are summarized as follows.
\begin{enumerate}
    \item The input to the GAN is (real) RF data that is highly complex and dynamic, subject to noise, channel, traffic, and interference effects. Statistical modeling may not be effective to represent the spectrum data. Therefore, the complex structures of deep neural networks in the GAN formulation can be effectively used to represent the spectrum data \cite{ROW19, PWN20, CHHM21}.
    \item The format of the RF data is not universal (like pixels in computer vision) and may vary from time series of in-phase/quadrature components (I/Qs) or
    Received Signal Strength Indicators (RSSIs) to frequency domain representations.
    \item The generator and the discriminator of the GAN may need to be distributed at transmitter and receiver, respectively, as data transmission and the corresponding reception need to be separated through a wireless channel in online applications. In this context, the message exchange between the generator and the discriminator may be executed over noisy wireless channels.
    \item While training the GAN, the RF data itself as well as communication channels between the generator and the discriminator may change due to the limited coherence time of channel conditions \cite{SDS19, SDS21}.
\end{enumerate}

Contributions of the paper are as follows. Sec.~\ref{sec:gans} introduces the basics of GANs and a number of alternative GAN structures, together with their underlying mathematics. GAN structures were originally developed for computer vision and image processing. This paper is on adopting them for applications in NextG wireless communications. To that end, Sec.~\ref{sec:proposed} first discusses performance measures used in computer vision and image processing. It introduces performance measures that can be used in wireless applications since the former measures cannot be used in this field. It introduces public datasets that can be employed in wireless applications. It also discusses performance measures for general classifiers, including wireless classifiers. It discusses the use of GANs to extend a given dataset, called {\em data augmentation.} Furthermore, Sec.~\ref{sec:proposed} introduces a number of GAN techniques that have been used in spectrum sharing, anomaly detection, and security for NextG wireless. A summary of these is available in Table 7. Sec.~\ref{sec:initres} presents an anomaly detection problem for the modulation classification task and presents our results where we solve this problem using two unsupervised deep learning methods, a GAN-based and an autoencoder-based algorithm. We also provide interpretations of the results, tradeoffs between the two approaches, and implementation details.
Sec.~\ref{sec:GANs4Wireless} provides a discussion of a number of potential future research directions, including their challenges. Finally, Sec.~\ref{sec:conclusion} provides our conclusions.

\section{Preliminaries on Generative Adversarial Networks}\label{sec:gans}
\ifCLASSOPTIONonecolumn
\begin{wrapfigure}{r}{0.55\textwidth}
%\vspace{-3mm}
\centering
\includegraphics[width=0.5\textwidth]{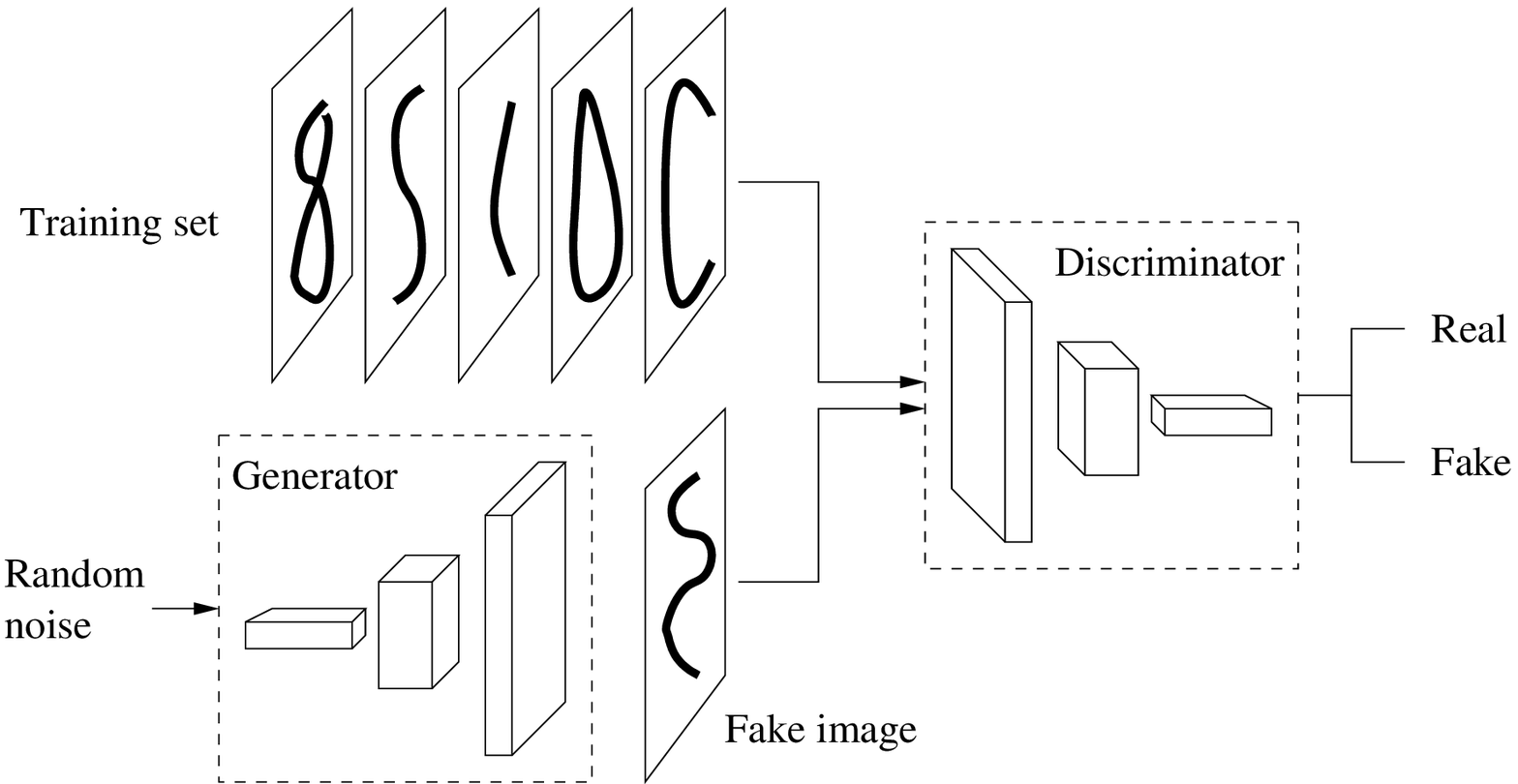}
\caption{Generative Adversarial Network (GAN) with image generation application \cite{Silva18}.}
\label{fig:GAN}
\end{wrapfigure}
\else
\vspace{3mm}
\begin{figure}[!t]
\centering
\includegraphics[width=0.48\textwidth]{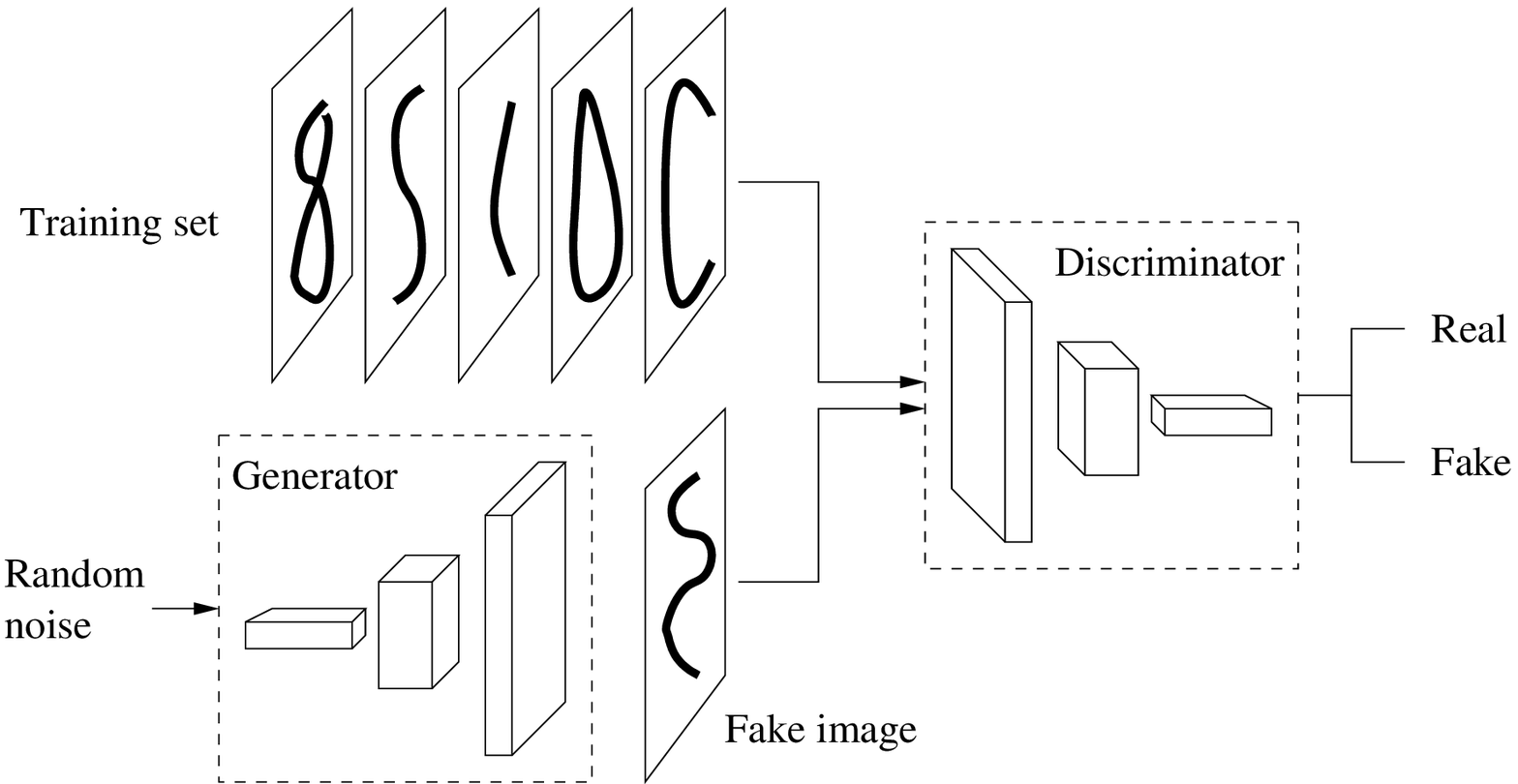}
\caption{Generative Adversarial Network (GAN) with image generation application \cite{Silva18}.}
\label{fig:GAN}
\end{figure}
\fi
GANs were introduced in 2014 \cite{Silva18,GPMXWOCB14}, and belong to the class of deep-learning-based generative models. Generative modeling is an unsupervised ML task that involves automatically discovering the regularities or patterns in input data in such a way that the model can be used to generate new samples that cannot be discriminated from the original dataset. Facebook's Director of Artificial Intelligence Research Yann Le Cunn is famously known to have said in June 2016 that ``GANs and the variations that are now being proposed is the most interesting idea in the last 10 years in ML, in my opinion.'' Given the many subsequent applications of GANs in different fields, some of which we discuss below, it is difficult to argue with this opinion.

GANs train a generative model by decomposing the problem into two sub-models, as depicted in Fig.~\ref{fig:GAN}. The first is a generator network trained to generate new examples, while the second is a discriminator network that tries to classify examples as either real (from the domain) or fake (generated). The two models are jointly trained that continues until the generator network learns to synthesize plausible samples that cannot be distinguished from real data.
%In other words, GANs are based on a game theoretic scenario in which the generator network must compete against an adversary. The generator network directly produces samples. Its adversary, the discriminator network, attempts to distinguish between samples drawn from the training data and samples drawn from the generator.

GANs and their variations have been used in a variety of applications. A few examples include creation of synthetic human faces \cite{KALL17,Kan17,KALL18} and transforming images from one domain (e.g., real scenery) to another domain (paintings by famous painters) \cite{ZPIE17}. Other examples of image-to-image translation in \cite{ZPIE17} include conversion of horse pictures to pictures of zebras and vice versa, pictures of summer scenery to pictures of the same scenery in winter, etc. Further applications include face aging \cite{ABD17}, creation of super-resolution images from those of low resolution \cite{Ledig17}, music generation \cite{YCY17}, audio synthesis \cite{DMP19}, and video synthesis \cite{Wang18}. More relevant to this paper, GANs have been applied to problems in spectrum sensing and security in wireless networks, e.g., \cite{DS18,TTN18,RMCP19,SDS19,UALQA19,YL19,ROW19}, and these will be discussed in more detail below. A list of the many different applications of GANs is available in \cite{GANZoo}.

To describe how GANs are trained, let {\bf x} belong to the manifold of the real input data with distribution $p_{\rm data} ({\bf x})$, and let {\bf z} belong to the latent or noise prior space with distribution $p_{\bf Z} ({\bf z})$. Further, let $G$ be a differentiable function representing the generator with input {\bf z}, and let $D$ be a differentiable function representing the discriminator with input {\bf x} or $G({\bf z})$, where the output of $D$ is mapped to the interval $[0, 1]$. Now consider the function
\ifCLASSOPTIONonecolumn
\begin{equation}
V (D, G) = \mathbb{E}_{{\bf x}\sim p_{\rm data} ({\bf x})} [\log D({\bf x})] + \mathbb{E}_{{\bf z}\sim p_{\bf Z} ({\bf z})} [\log (1 - D(G({\bf z})))]\label{eqn:V(D,G)} \; .
\end{equation}
\else
\begin{equation}
\begin{split}
V (D, G)  = & \mathbb{E}_{{\bf x}\sim p_{\rm data} ({\bf x})} [\log D({\bf x})]\label{eqn:V(D,G)}\\
            & + \mathbb{E}_{{\bf z}\sim p_{\bf Z} ({\bf z})} [\log (1 - D(G({\bf z})))] \; .
\end{split}
\end{equation}
\fi
The first term increases when the real samples are more correctly classified, while the second term increases as the discriminator more successfully identifies the generated samples as fake. Thus, the discriminator $D$ acts to maximize~\eqref{eqn:V(D,G)}. On the other hand, $G$ attempts to minimize~\eqref{eqn:V(D,G)}, that is equivalent to minimizing just the second term, which increases the likelihood that the discriminator is fooled. This leads to a minimax optimization in which $D$ and $G$ work against each other to achieve the equivalent of a Nash equilibrium point \cite{GPMXWOCB14,Goodfellow16}:
\begin{equation}
\min_G\max_D V(D,G) \; .
\label{eqn:minmax}
\end{equation}
The two optimizations in~(\ref{eqn:minmax}) are carried out by employing neural networks and backpropagation via gradient ascent and gradient descent, which is possible since $D$ and $G$ are differentiable.

Given a batch $\{{\bf x}_i, {\bf z}_i\}_{i=1}^n$ of training data and samples from the latent space, we can convert the optimization expressed via~(\ref{eqn:V(D,G)}) and~(\ref{eqn:minmax}) into the optimization of two cost functions, for $D$ and $G$, respectively, as
\ifCLASSOPTIONonecolumn
\begin{equation}
J_D = -\frac{1}{n} \bigg(\sum_{i=1}^n \log D({\bf x}_i) + \sum_{i=1}^n \log (1-D(G({\bf z}_i))\bigg),\qquad J_G = -\frac{1}{n} \sum_{i=1}^n \log \big( 1 - D(G({\bf z}_i))\big) .\label{eqn:J_D-J_G}
\end{equation}
\else
\begin{equation}
\begin{split}
J_D & = -\frac{1}{2} \big
(\sum_{i=1}^n \log D({\bf x}_i) + \sum_{i=1}^n \log (1-D(G({\bf z}_i))\big
), \label{eqn:J_D-J_G} \\
J_G & = -\frac{1}{n} \sum_{i=1}^n \log D(G({\bf z}_i)).
\end{split}
\end{equation}
\fi
Details on implementing~(\ref{eqn:J_D-J_G}) with a neural network and gradient ascent and descent to train the GAN can be found in \cite[Algorithm 1]{GPMXWOCB14}.
Note that by substituting the criterion in (\ref{eqn:V(D,G)})--(\ref{eqn:minmax}) with that in (\ref{eqn:J_D-J_G}), an implicit assumption of ergodicity, or ensemble averages being equal to time averages, is made. This assumption, common in signal processing and communications, will be made again in the sequel.

Additional information that is correlated with the input data, such as class labels, can be used to improve GAN performance, either in the form of more stable or faster training, or generated images that have better quality. Such conditional GANs (CGANs) are trained in such a way that both the generator and the discriminator models are conditioned on the class label, so that when the trained generator is used as a standalone model to generate samples in the domain, samples of a given type can be generated \cite{MO14}. For example, in the synthesis of faces one could focus on generating a female face, or one could convert a summer scene into a winter scene, etc. Popular face aging applications are also based on this principle. To describe this approach, let $y$ represent extra information on which the generator and discriminator are conditioned. One can perform the conditioning by feeding $y$ into both the discriminator and generator as additional input layer. As mentioned above, the information $y$ could be in the form of a class label, leading to a so-called class-conditional GAN, or some other kind of input such as a single image, in the case the GAN is performing an image-to-image translation task. Letting ${\bf y}$ represent a potentially multidimensional label from a distribution $p_{\bf Y}({\bf y})$, the optimization in (\ref{eqn:V(D,G)}) and (\ref{eqn:minmax}) becomes
\ifCLASSOPTIONonecolumn
\begin{equation}
\min_G\max_D\mathbb{E}_{{\bf x},{\bf y}\sim p_{\rm data} ({\bf x},{\bf y})} [\log D({\bf x},{\bf y})] + \mathbb{E}_{{\bf y}\sim p_{\bf Y}({\bf y}), {\bf z}\sim p_{\bf Z} ({\bf z})} [\log (1 - D(G({\bf z, {\bf y}), {\bf y}}))]
\label{eqn:CGANminmax}
\end{equation}
\else
\begin{equation}
\begin{split}
\min_G\max_D\quad & \mathbb{E}_{{\bf x},{\bf y}\sim p_{\rm data} ({\bf x},{\bf y})} [\log D({\bf x},{\bf y})] \label{eqn:CGANminmax}\\
& + \mathbb{E}_{{\bf y}\sim p_{\bf Y}({\bf y}), {\bf z}\sim p_{\bf Z} ({\bf z})} [\log (1 - D(G({\bf z, {\bf y}), {\bf y}}))]
\end{split}
\end{equation}
\fi
for CGANs, with (\ref{eqn:J_D-J_G}) updated in a straightforward fashion based on~(\ref{eqn:CGANminmax}).

A number of other GAN formulations exist beyond those mentioned above. For example, in the Least-Squares GAN (LSGAN) \cite{MLXLWS17}, the objective functions become
\ifCLASSOPTIONonecolumn
\begin{equation}
\min_D \{ \mathbb{E}_{{\bf x}\sim p_{\rm data}({\bf x})} [( D ({\bf x}) - a )^2 ] + \mathbb{E}_{{\bf z}\sim p_{\bf Z}({\bf z})} [(D(G({\bf z}))-b)^2] \},\quad
\min_G \mathbb{E}_{{\bf z}\sim p_{\bf Z}({\bf z})}[(D(G({\bf z}))-c)^2] ,
\end{equation}
\else
\begin{equation}
\begin{split}
\min_D & \{ \mathbb{E}_{{\bf x}\sim p_{\rm data}({\bf x})} [( D ({\bf x}) - a )^2 ] \\
       & + \mathbb{E}_{{\bf z}\sim p_{\bf Z}({\bf z})} [(D(G({\bf z}))-b)^2] \},\\
\min_G & \mathbb{E}_{{\bf z}\sim p_{\bf Z}({\bf z})}[(D(G({\bf z}))-c)^2] ,
\end{split}
\end{equation}
\fi
where $a$ and $b$ represent labels for real data and fake data, respectively, and $c$ be the value that $G$ wants $D$ to believe for fake data. Reference \cite{YL19} employs an LSGAN with $a = - b = 1$ and $c=0$. Another variation is known as Wasserstein GAN (WGAN) \cite{ACB17}, which employs the loss function
\begin{equation}
\min_G\max_{D\in {\cal D}}\mathbb{E}_{{\bf x}\sim p_{\rm data}({\bf x})}[D({\bf x})] -\mathbb{E}_{{\bf z}\sim p_{\bf Z}({\bf z})}[D(G({\bf z}))],
\end{equation}
where ${\cal D}$ is a set referred to as the set of 1-Lipshitz functions. Reference \cite{GAADC17} removes the condition $D\in {\cal D}$ but adds an additional term for stability, repeatability, and predictable behavior during the training process. Reference \cite{ROW19} employs a WGAN with the additional term proposed in \cite{GAADC17}. An alternative way to enforce the Lipshitz constraint is to add a penalty on the gradient norm for random samples ${\hat{\textbf{x}}}\sim p_{\hat X}$, and the updated loss function can be expressed as
\ifCLASSOPTIONonecolumn
\begin{equation}
\min_G\max_{D\in {\cal D}} \mathbb{E}_{{\bf x}\sim p_{\rm data}({\bf x})}[D({\bf x})] -\mathbb{E}_{{\bf z}\sim p_{\bf Z}({\bf z})}[D(G({\bf z}))] + \lambda \mathbb{E}_{{\hat{\textbf{x}}}\sim p_{\hat X}(\hat{\bf x})}\left[ \left( \| \nabla_{\hat{\textbf{x}}} D(\hat{\textbf{x}})\|_2  -1 \right)^2 \right].
\end{equation}
\else
\begin{equation}
\begin{split}
\min_G\max_{D\in {\cal D}} \quad & \mathbb{E}_{{\bf x}\sim p_{\rm data}({\bf x})}[D({\bf x})] -\mathbb{E}_{{\bf z}\sim p_{\bf Z}({\bf z})}[D(G({\bf z}))] \\
 &+ \lambda \mathbb{E}_{{\hat{\textbf{x}}}\sim p_{\hat X}(\hat{\bf x})}\left[ \left( \| \nabla_{\hat{\textbf{x}}} D(\hat{\textbf{x}})\|_2  -1 \right)^2 \right].
\end{split}
\end{equation}
\fi
This approach is known as WGAN with gradient penalty (WGAN-GP) \cite{GAADC17}  and it provides stability and predictable behavior during the GAN training process.
\subsection{Anomaly Detection by Using GANs}\label{sec:anomalydetection}
One area of particular importance where GANs have been applied is in anomaly detection, which is the task of discovering anomalies, or patterns in the data that do not conform to ``normal behavior.''  While the use of GANs is based on modeling normal behavior using the adversarial training process, an {\em  anomaly score} based on this model can be calculated and employed for anomaly detection. Several GAN-based approaches for anomaly detection have been published, all of which employ the technique of Adversarial Feature Learning idea \cite{DKD16}. This idea makes use of a novel architecture referred to as bidirectional GANs, or BiGANs.

\begin{figure}[!t]
\centering
\ifCLASSOPTIONonecolumn
\vspace{-3mm}
\scalebox{0.65}{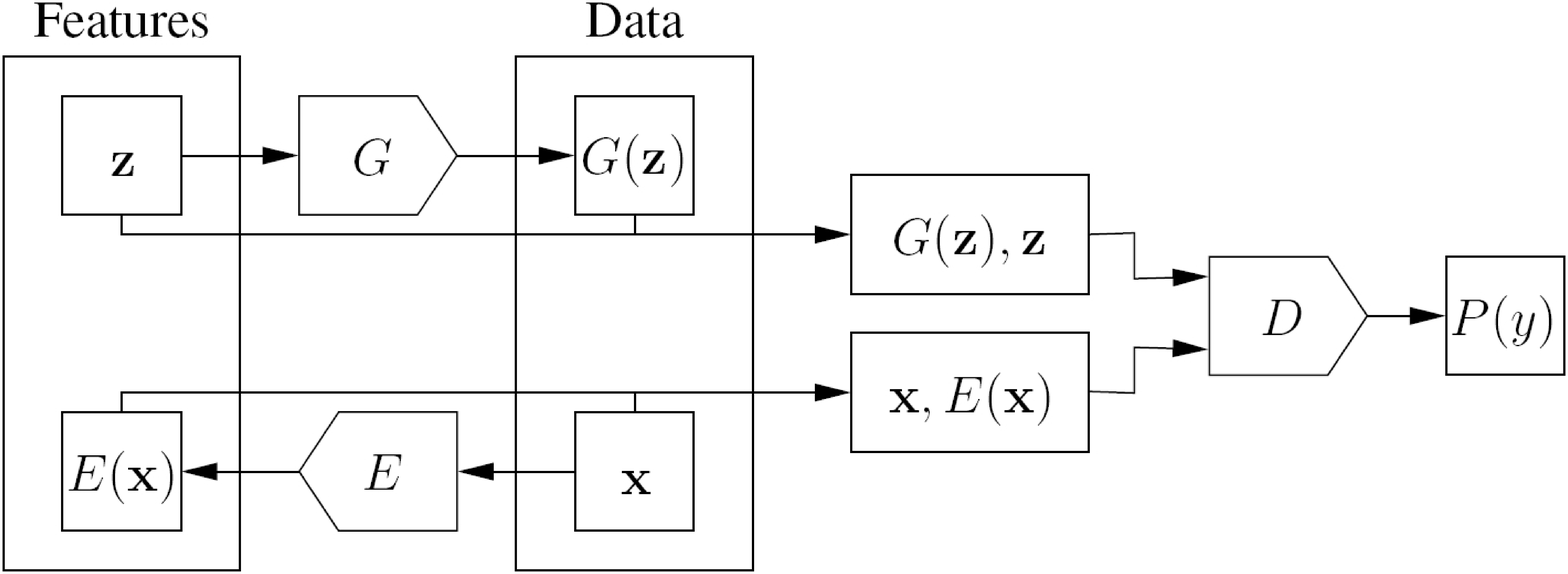}
\else
\vspace{1mm}
\includegraphics[width=0.48\textwidth]{Figures/BiGAN.eps}
\fi
\caption{The structure of BiGAN \cite{DKD16}.}
\label{fig:BiGAN}
\end{figure}
BiGANs add an inverse mapping from the data space to the latent distribution, from which regular GANs generate artificial samples. The overall model is depicted in Fig.~\ref{fig:BiGAN}. A BiGAN includes an encoder $E$ which maps data {\bf x} to latent representations {\bf z}, in addition to the generator $G$ employed by the standard GAN architecture \cite{DKD16}. The BiGAN discriminator $D$ discriminates not only in data space ({\bf x} versus $G({\bf z})$), but jointly in both the data and latent space (tuples $({\bf x};E({\bf x}))$ versus $(G({\bf z}); {\bf z}))$, where the latent component is either an encoder output $E({\bf x})$ or a generator input ${\bf z}$.

The two modules $E$ and $G$ do not directly ``communicate'' with one another: the encoder never ``sees'' generator outputs ($E(G({\bf z}))$ is not computed), and vice versa. However, it is shown in \cite{DKD16} that the encoder and generator must learn to invert one another in order to fool the BiGAN discriminator. In other words, $E$ learns the inverse of the generator $E=G^{-1}$. The encoder $E$ is a nonlinear parametric function in the same way as $G$ and $D$, and can be trained using a standard learning algorithm such as gradient descent. A latent representation {\bf z} may be thought of as a ``label'' for {\bf x}, but one which came for ``free,'' without the need for supervision. As discussed in \cite{DKD16}, BiGANs make no assumptions about the structure or type of data to which they are applied.

In this paper, we will focus on three GAN architectures that conceptually employ the inverse generator concept of BiGANs. These are the AnoGAN \cite{SSWSL17}, Efficient GAN-Based Anomaly Detection (EGBAD) \cite{ZFLMC18}, and a GAN + autoencoder approach \cite{AAB18}. AnoGAN is a deep convolutional GAN that learns a manifold of normal variability, together with a scoring scheme that labels anomalies based on the mapping from the image to the latent space. For results on medical imaging data, see \cite{SSWSL17}. Given a query image ${\bf x}$, the algorithm iterates through points in the latent space to find a representation $G({\bf z})$ that is close to ${\bf x}$. The algorithm begins by choosing a random point ${\bf z}_1$ in the latent space, and generates a data sample $G({\bf z}_1)$. The algorithm then proceeds through a number of points ${\bf z}_1, {\bf z}_2, \ldots, {\bf z}_\Gamma$ based on the following loss function:
\begin{equation}
{\cal L}({\bf z}_\gamma) = (1-\lambda ) {\cal L}_R({\bf z}_\gamma ) + \lambda {\cal L}_D ({\bf z}_\gamma) \; , \label{eqn:L}
\end{equation}
where
\begin{equation}
{\cal L}_R({\bf z}_\gamma)=\| {\bf x} - G({\bf z}_\gamma)\|_1\label{eqn:L_R}
\end{equation}
is the residual loss, and
\begin{equation}
{\cal L}_D({\bf z}_\gamma) = \| f({\bf x}) - f(G({\bf z}_\gamma))\|_1\label{eqn:L_D}
\end{equation}
is the discriminator loss, where $f$ is the output of one of the layers of the multilayer perceptron and $0<\lambda < 1$ is an interpolation coefficient. The residual loss enforces similarity between the query data ${\bf x}$ and the generated data $G({\bf z}_\lambda)$, while the discrimination loss constrains the generated data to lie near the learned manifold ${\cal X}$ \cite{SGZCRC16}. For each $\gamma =  2, 3, \ldots, \Gamma$, ${\bf z}_{\gamma}$ is calculated by iteratively minimizing (\ref{eqn:L}) via backpropagation steps. The iteration is on ${\bf z}_\gamma$, the coefficients of $G$ and $D$ are not changed. The approach of using a linear combination of two loss functions (\ref{eqn:L_R}) and (\ref{eqn:L_D}) that employ the $L_1$ norm for training a deep neural network is different than the approaches discussed earlier for training GANs and CGANs. This approach originated with \cite{YCLSHD17} and was adapted in \cite{SSWSL17}.

Implementing the AnoGAN optimization over $\Gamma$ steps results in a relatively high computational load. The EGBAD approach \cite{ZFLMC18} was developed as a more efficient alternative and is based on the methods in \cite{DKD16,DBPLAMC17}, which enable learning an encoder $E$ by mapping input samples to their latent representation during adversarial training. GANomaly \cite{AAB18} is designed to be an improvement over \cite{DKD16,SSWSL17,ZFLMC18} in terms of both performance and speed. The algorithm trains a generator network to learn the manifold of the input samples while at the same time training an autoencoder to encode the data in their latent representation. This approach uses a discriminator, a decoder, and two encoders, but the encoders have the same architecture. We describe this approach in more detail below.

\begin{figure}[!t]
\centering
\ifCLASSOPTIONonecolumn
\vspace{-3mm}
\scalebox{0.65}{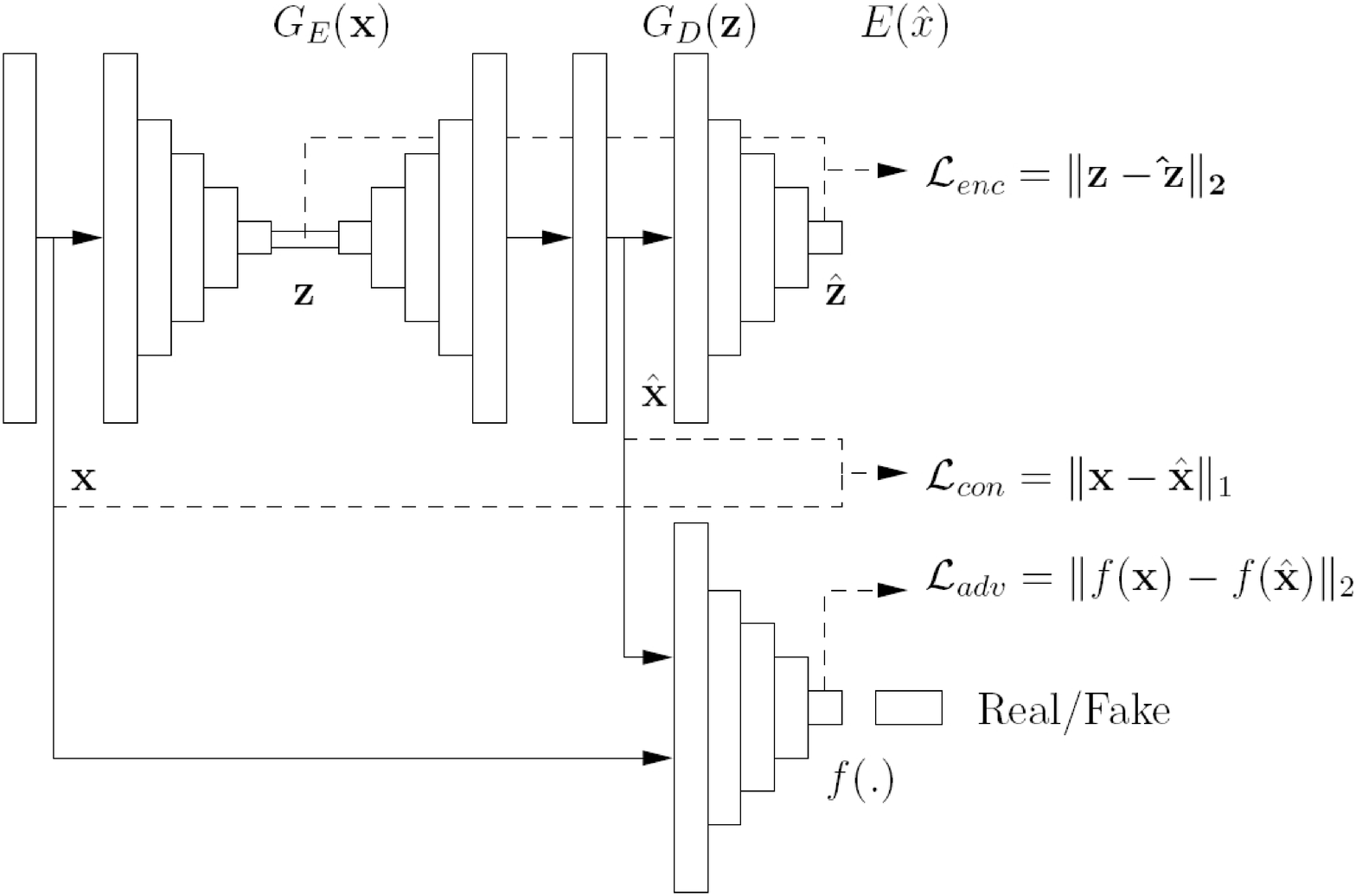\hspace{1.2in}}
\else
\vspace{1mm}
\includegraphics[width=0.48\textwidth]{Figures/GANomaly.eps}
\fi
\caption{Simplified single-dimensional architecture representation of GANomaly. Multiple blocks in series represent two-dimensional convolutional encoders and single blocks are for providing two-dimensional input-output \cite{AAB18}.}
\label{fig:GANomaly}
\end{figure}

\ifCLASSOPTIONonecolumn
\begin{figure}[!b]
\centering
\scalebox{0.65}{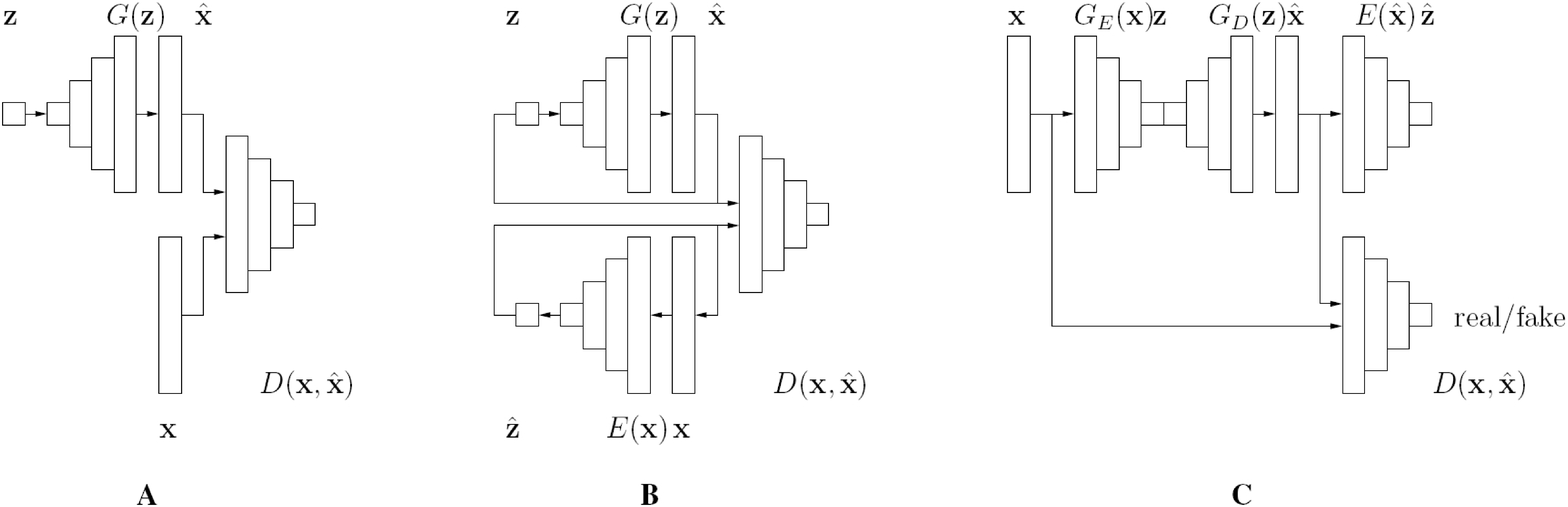\hspace{1in}}
\caption{Comparison of GANs for anomaly detection for computer vision and image processing applications, A: AnoGAN \cite{SSWSL17}, B: EGBAD \cite{ZFLMC18}, C: GANomaly \cite{AAB18}. Two-dimensional convolutional encoders and input-output devices are depicted as single-dimensional blocks for simplicity without any loss in functionality.}
\label{fig:AnomalyGAN}
\end{figure}
\else
\begin{figure*}[!t]
\centering
\includegraphics[width=0.96\textwidth]{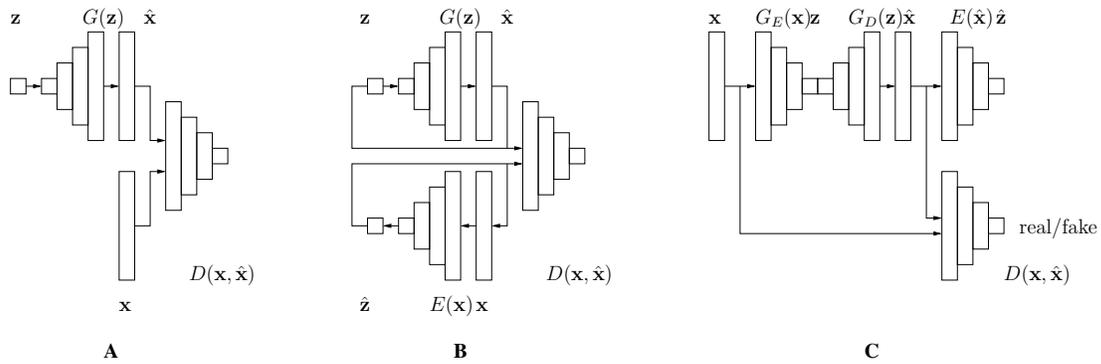}
\caption{Comparison of GANs for anomaly detection for computer vision and image processing applications, A: AnoGAN \cite{SSWSL17}, B: EGBAD \cite{ZFLMC18}, C: GANomaly \cite{AAB18}. Two-dimensional convolutional encoders and input-output devices are depicted as single-dimensional blocks for simplicity without any loss in functionality.}
\label{fig:AnomalyGAN}
%\end{wrapfigure}
\end{figure*}
\fi

As shown in Fig.~\ref{fig:GANomaly}--\ref{fig:AnomalyGAN}, the generator network has three elements in series: an encoder $G_E$, a decoder $G_D$, and another encoder $E$.  The combination of $G_E$ and $G_D$ forms an autoencoder that functions as the generator $G$. The encoder $E$ has the same structure as $G_E$. $G_E$ takes the data sample ${\bf x}$ and generates an encoded version ${\bf z}$ in the latent space. Then $G_D$ employs ${\bf z}$ to create $\hat{\bf x}$, which is a reconstructed form of ${\bf x}$. Finally, $\hat{\bf x}$ is used to generate another point in the latent space, $\hat{\bf z}$. Three loss functions are defined, which are combined to generate the overall generator loss. The first is the adversarial loss ${\cal L}_{\rm adv}$,
\ifCLASSOPTIONonecolumn
\begin{equation}
{\cal L}_{\rm adv}=\mathbb{E}_{{\bf x}\sim p_{\bf X}({\bf x})} \| f({\bf x}) - \mathbb{E}_{{\bf x}\sim p_{\bf X} ({\bf x})} f (G({\bf x})) \|_2 = \|f({\bf x})-f(\hat{\bf x})\|_2 \; , \label{eqn:Ladv}
\end{equation}
\else
\begin{equation}
\begin{split}
{\cal L}_{\rm adv} &=\mathbb{E}_{{\bf x}\sim p_{\bf X}({\bf x})} \| f({\bf x}) - \mathbb{E}_{{\bf x}\sim p_{\bf X} ({\bf x})} f (G({\bf x})) \|_2 \\  \label{eqn:Ladv}
&= \|f({\bf x})-f(\hat{\bf x})\|_2 \; ,
\end{split}
\end{equation}
\fi
where $f$ is the output of an intermediate layer in the multi-layer perceptron.
The second loss function is the contextual loss and is given by
\begin{equation}
{\cal L}_{\rm con}=\mathbb{E}_{{\bf x}\sim p_{\bf X}}\| {\bf x} - G({\bf x}) \|_1 = \|{\bf x}-\hat{\bf x}\|_1 \; , \label{eqn:Lcon}
\end{equation}
and the third is the encoder loss
\begin{equation}
{\cal L}_{\rm enc} = \mathbb{E}_{{\bf x}\sim p_{\bf X}({\bf x})} \| G_E({\bf x}) - E(G({\bf x})) \|_2 = \| {\bf z}-\hat{\bf z}\|_2.\label{eqn:Lenc}
\end{equation}
Finally, the generator loss is given as
\begin{equation}
{\cal L} = w_{\rm adv} {\cal L}_{\rm adv} +  w_{\rm con} {\cal L}_{\rm con} +  w_{\rm enc} {\cal L}_{\rm enc} \; , \label{eqn:L-GANomaly}
\end{equation}
where $w_{\rm adv}$, $w_{\rm con}$, and $w_{\rm enc}$ are coefficients between 0 and 1 that sum to 1. During the test stage, the model uses ${\cal L}_{\rm enc}$ given in (\ref{eqn:Lenc}) for scoring the abnormality $A(\hat{\bf x})$ of a given image:
\begin{equation}
A(\hat{\bf x}) = \| G_E(\hat{\bf x}) - E(G(\hat{\bf x})) \|_1 .\label{eqn:anomalyscore}
\end{equation}

In order to make the anomaly score easier to interpret, \cite{AAB18} proposes to compute it for every sample $\hat{\bf x}$ in the test set $\hat D$ to obtain the set $S = \{s_i : A(\hat{\bf x}, \hat{\bf x} \in \hat D \}$ of individual anomaly scores, and then apply a scaling to force the scores to lie within the range $[0, 1]$:
\begin{equation}
s_i^\prime = \frac{s_i-\min(S)}{\max(S)-\min(S)}.
\end{equation}

\ifCLASSOPTIONonecolumn
\begin{table}[!t]
\small
\centering
\begin{tabular}{|l|l|l|c|}
\hline
Abbreviation & Application & Comment & Reference \\
\hline
GAN & General &Employs (1)-(3). Structure in Fig.~1.& [6], [14]\\                     %\cite{Silva18,GPMXW0CB14} \\
CGAN & &Conditional GAN. Employs (4). &\cite{MO14}\\
LSGAN & &Least Squares GAN. Employs (5). &\cite{MLXLWS17}\\
WGAN & &Wasserstein GAN. Employs (6). &\cite{ACB17}\\
WGAN-GP & &Gradient Penalty WGAN. Employs (7). &\cite{GAADC17}\\
\hline
BiGAN & Anomaly Detection &Bidirectional GAN. Structure in Fig.~2. &\cite{DKD16}\\
AnoGAN & &Anomaly GAN. Employs (8)-(10).& \cite{SSWSL17} \\
EGBAD & &Efficient GAN-Based Anomaly Detection. &\cite{ZFLMC18} \\
GANomaly & &GAN+Autoencoder. Employs (11)-(16). Structure in Fig.~3. &\cite{AAB18} \\
\hline
\end{tabular}
\caption{Different versions of GANs used in computer vision and image processing applications. Note that Fig.~\ref{fig:AnomalyGAN} compares the structures of AnoGAN, EGBAD, and GANomaly. Which of GAN, CGAN, LSGAN, WGAN, or WGAN-GP is preferable depends on the application. For the tradeoffs among the versions used in anomaly detection, see the text.}
\label{tbl:GANs-CVIP}
\end{table}
\else
\begin{table*}[!t]
\small
\centering
\begin{tabular}{|l|l|l|c|}
\hline
Abbreviation & Application & Comment & Reference \\
\hline
GAN & General &Employs (1)-(3). Structure in Fig.~1.& [6], [14]\\                     %\cite{Silva18,GPMXW0CB14} \\
CGAN & &Conditional GAN. Employs (4). &\cite{MO14}\\
LSGAN & &Least Squares GAN. Employs (5). &\cite{MLXLWS17}\\
WGAN & &Wasserstein GAN. Employs (6). &\cite{ACB17}\\
WGAN-GP & &Gradient Penalty WGAN. Employs (7). &\cite{GAADC17}\\
\hline
BiGAN & Anomaly Detection &Bidirectional GAN. Structure in Fig.~2. &\cite{DKD16}\\
AnoGAN & &Anomaly GAN. Employs (8)-(10).& \cite{SSWSL17} \\
EGBAD & &Efficient GAN-Based Anomaly Detection. &\cite{ZFLMC18} \\
GANomaly & &GAN+Autoencoder. Employs (11)-(16). Structure in Fig.~3. &\cite{AAB18} \\
\hline
\end{tabular}
\caption{Different versions of GANs used in computer vision and image processing applications. Note that Fig.~\ref{fig:AnomalyGAN} compares the structures of AnoGAN, EGBAD, and GANomaly. Which of GAN, CGAN, LSGAN, WGAN, or WGAN-GP is preferable depends on the application. For the tradeoffs among the versions used in anomaly detection, see the text.}
\label{tbl:GANs-CVIP}
\end{table*}
\fi

Fig.~\ref{fig:AnomalyGAN} shows a comparison of the architectures of the three GAN-based anomaly detection algorithms discussed above.
%In this figure, A stands for AnoGAN \cite{SSWSL17}, B for EGBAD \cite{ZFLMC18}, and C for GANomaly \cite{AAB18}.
Experiments show that EGBAD is faster than AnoGAN, and that GANomaly achieves better performance and speed than EGBAD \cite{AAB18,MGSG19}. Table~\ref{tbl:GANs-CVIP} provides a summary of GAN versions used for computer vision and image processing applications.

\section{Application of GANs to 5G and Beyond}\label{sec:proposed}
In what follows, we will first discuss, in Sec.~\ref{sec:perf_meas_wireless}, measures of performance and specific issues that have to do with applying GANs first to computer vision and image processing and then to wireless applications. We will discuss public datasets available for enabling the training and benchmarking the performance of GANs in the wireless domain. In the same section, we will discuss measures for general classifier performance. We will specifically use these measures in wireless classifier applications. In Sec.~\ref{sec:augmentation}, we will discuss the technique of data augmentation by using GANs, which can be employed in the following three sections. Then, in Sec.~\ref{sec:sharing}--Sec.~\ref{sec:security}, we will discuss
fast, accurate, and robust methods for analyzing large quantities of spectrum data in order to identify opportunities for {\em i)} spectrum sharing, {\em ii)} detecting anomalies, and {\em iii)}
mitigating security attacks, respectively.
\subsection{Performance Measures for GANs and General Classifiers}\label{sec:perf_meas_wireless}
We will now discuss performance measures for GANs, first for computer vision and image processing, and then for wireless applications, and finally, general classifiers.
There are basically two measures of the quality of images generated by
GANs for computer vision and image processing applications. These are
Inception Score (IS) \cite{SGZCRC16} and Fr\'echet Inception Distance
(FID) \cite{DL82,HRUNH18}. IS offers a way to objectively and
quantitatively evaluate the quality of generated images by a GAN.
It is generally considered that this score is
well-correlated with scores from human observers. Similarly, FID is
used to evaluate the quality of images generated by GANs. FID
compares the mean and standard deviation of one of the deeper layers
in a Convolutional Neural Network (CNN) named Inception-v3. These layers are
closer to output nodes and are believed to mimic human perception of
similarity in images. Clearly, these two measures are useless for wireless applications.
They need to be replaced with measures meaningful in the context of wireless communications.

For wireless applications, the goal is to calculate the
similarity of two probability densities. A commonly used measure
towards that end is the Jensen-Shannon distance, which is the square
root of the Jensen-Shannon divergence \cite{JSD}. Jensen-Shannon divergence, on
the other hand, is a symmetric form of the well-known Kullback-Leibler
divergence \cite{CT06}.

In passing, we would like to state that to facilitate the training of GANs in the wireless domain and benchmark the performance of GANs, there are increasingly more public datasets available.
Examples of such sets are
RADIOML 2016.04C, RADIOML 2016.10A, RADIOML 2018.01A by DeepSig \cite{DeepSig}, RFMLS 2016a by DARPA \cite{RFMLS}, CBRS by NIST \cite{NIST}, and synthetic data generated by GNU Radio \cite{OW16}.

To measure the performance of any classifier, Probability of Detection ($P_D$) and Probability of False Alarm ($P_{FA}$) are calculated. Then, the Receiver Operating Characteristics (ROC), which plot $P_D$ against $P_{FA}$ are drawn \cite{ROC}. A commonly used measure is Area Under Curve of the ROC (AUROC). Accuracy (ACC) is defined as the ratio of the correct classifications to the total classifications. It can be calculated as Accuracy = (True Positives + True Negatives)/ (True Positives + False Positives + True Negatives + False Negatives). Furthermore, the variables Precision, Recall, and F1 Score determine the success of a binary classification problem. Precision equals the ratio of the true positives to total (true and false) classified positives. Recall equals the number of true positives to the sum of true positives and false negatives. Precision is a good measure to employ when the cost of the false positive is high whereas Recall is a good measure when the cost of the false negative is high. F1 Score, the harmonic mean \cite{HarmonicMean} of
Precision and Recall \cite{F1-score}, is employed when it is desired to have a balance between Precision and Recall. All three measures take values between 0 and 1. For a given classification system, it is desirable that each be as close to 1 as possible. Please see Sec.~\ref{sec:initres} for combinations of these quantities and the definitions of Density and Coverage in \cite{NOUCY20}.

An important lesson learned in this subsection is that while it is desirable to employ GAN structures from computer vision and image processing for NextG wireless applications, the performance measures in the two fields are different.
\subsection{Data Augmentation via GANs for NextG}\label{sec:augmentation}
By recognizing that supervised ML requires significant number of training data samples and it is expensive and likely even infeasible to collect a sufficiently comprehensive and representative set of training data,
\cite{DS18} provides {\em training data augmentation\/} by adding synthetic data to an existing training set. This is achieved by a workflow of three steps. In the first step, a CGAN is trained using real training samples. In this step, the generator learns to synthesize new data samples. In the second step, the synthetic data samples are used along with the real samples to train a classifier. In the third step, as new data comes in, the classifier is used for classification purposes. The authors of \cite{DS18} show that this approach significantly improves the adaptation time and accuracy of the resulting spectrum sensing.

In an attempt to come up with a good modulation recognition technique, \cite{TTZL18} employs Auxiliary Classifier GANs (AC-GANs) \cite{OOS16} after a density transformation of the signal called Contour Stellar Image to enhance the performance of CNNs. Although employing a training sequence whose length is 10\% of another algorithm, \cite{TTZL18} achieves up to 6\% gain in the recognition performance of the test sequence.

Reference \cite{CHHM21} investigates training data augmentation for deep learning RF systems by asking basic questions about the augmentation process. The paper concentrates on the Automatic Modulation Classification (AMC) problem. These questions are: {\em i)\/} how useful a synthetically trained system will be when deployed without considering the environment within the synthesis, {\em ii)\/} how can augmentation be leveraged within the RF ML domain, and {\em iii)\/} what impact knowledge of degradations to the signal caused by the transmission channel contributes to the performance of a system. The results show that for data generation to a higher fidelity, the propagation path from the Digital-to-Analog Converter (DAC) to the Analog-to-Digital Converter (ADC) must be investigated and modeled. Second, although augmentation provides savings in terms of time and money, the authors suggest a cost analysis to achieve a balance between the two. Finally, a methodology is established for the quantity of data needed.

A lesson that should be derived here is that GANs can generate synthetic training data and thus augment the training set in ML. We will discuss this topic in more detail in Sec.~\ref{sec:GANs4Wireless}.
\subsection{Spectrum Sharing}\label{sec:sharing}
As NextG networks rely on the co-existence of heterogeneous networks such as cellular and non-terrestrial networks, and radar systems, spectrum sharing is expected to play a major role in NextG communications and radar systems,
while relying on ML techniques to identify and make use of spectrum opportunities. One particular example is Citizens Broadband Radio Service (CBRS) band at 3.5 GHz where the cellular wireless system needs to share the spectrum with the tactical radar while avoiding mutual interference \cite{caromi2018detection}. As such, the use of ML for spectrum sensing problems is currently a growing area of interest, as evidenced in the survey \cite{AK19} which summarizes results from \cite{GNAHP16,LZWF16,WY16,KZH17,DWYWC13,MGW14,FJB11,LP16,CSZHHSLW17,NDZW17}. The work reported in \cite{GNAHP16,LZWF16,WY16,KZH17} focuses on ML issues including optimization of supervised classifiers, reduction of the feature space dimension, reduction of interference, and minimization of the number of sensors, respectively. Attacker detection and minimization of sensing time, delay, and operations in collaborative spectrum sensing are discussed in \cite{DWYWC13,MGW14}. Techniques that target Spectrum Sensing Data Falsification (SSDF) are developed in \cite{FJB11,LP16,CSZHHSLW17,NDZW17}. Furthermore, \cite{TZL10,VPV12,HSZBWLG17,YHWL19} discuss the use of neural networks for spectrum sensing under low Signal-to-Noise Ratio (SNR) conditions. References \cite{CS19,YPLRGMC19} discuss employing neural networks for spectrum sensing in the face of parameter uncertainties in the transmitted signals.
%Since there is a limited amount of existing research in this area, it is open to the possibility of making a significant contribution, as sought by the solicitation.

Spectrum sharing techniques can be broadly characterized as belonging to open sharing or hierarchical
access models \cite{ZS07,Peha09,BJ16}. In the open sharing model, each network accesses the same spectrum without any interference constraint from one network to its peers. The unlicensed band is an example of this model. The hierarchical or cognitive radio model consists of a primary network and a secondary network that accesses the primary spectrum without interfering with the Primary Users (PUs). We will focus on the concept of cognitive radio networks in order to achieve efficient spectrum sharing via ML, using the classification from \cite{GJMS09} to organize potential approaches to spectrum sharing in cognitive radio networks. In this classification, spectrum sharing techniques are categorized as either underlay, overlay, or interweave.

\ifCLASSOPTIONonecolumn
\begin{table}[!t]
\else
\begin{table*}[!t]
\fi
\small
\begin{center}
\begin{tabular}{lll}
\hline
{\bf Underlay}                        &{\bf Overlay}                     &{\bf Interweave}                     \\
\hline
SU knows the channel strengths        & SU knows the channel gains,      & When PU is not using the spectrum,  \\
of PU                                 & codebooks, and messages of PU    & SU knows the spectral holes in      \\
                                      &                                  & space, time, or frequency           \\
\hline
As long as the interference is below  & SU can simultaneously transmit   & SU can simultaneously transmit      \\
an acceptable limit, SU can           & with PU; interference to PU      & with PU only in the case of false   \\
simultaneously transmit with PU       & can be offset by using part of   & spectral hole detection             \\
                                      & SU's power to relay PU's message &                                     \\
\hline
SU's transmit power is limited by     & SU can transmit at any power;    & SU's power is limited by the        \\
the interference constraint           & the interference to PU can be    & range of its spectral holes         \\
                                      & offset by relaying PU's message  &                                     \\
\hline
\end{tabular}
\end{center}
\caption{Characteristics of underlay, overlay, and interweave techniques for cognitive radio networks \cite{BJ16,GJMS09}.}
\label{tbl:cognet}
\ifCLASSOPTIONonecolumn
\end{table}
\else
\end{table*}
\fi
Table~\ref{tbl:cognet} summarizes the basic characteristics of underlay, overlay, and interweave techniques \cite{GJMS09}.
In the underlay technique, a Secondary User (SU) needs to be aware of the channels for all active PUs in order to know if the interference it generates will be below an acceptable limit. This consideration limits its transmit power. For the overlay technique, in addition to the PU channel characteristics, the SU also knows the codebooks and messages of the active PUs. The SU can transmit at
\ifCLASSOPTIONonecolumn
\begin{wrapfigure}{r}{0.5\textwidth}
\vspace{2mm}
\else
\begin{figure}{!b}
\vspace{-1mm}
\fi
\centering
\includegraphics[width=0.5\textwidth]{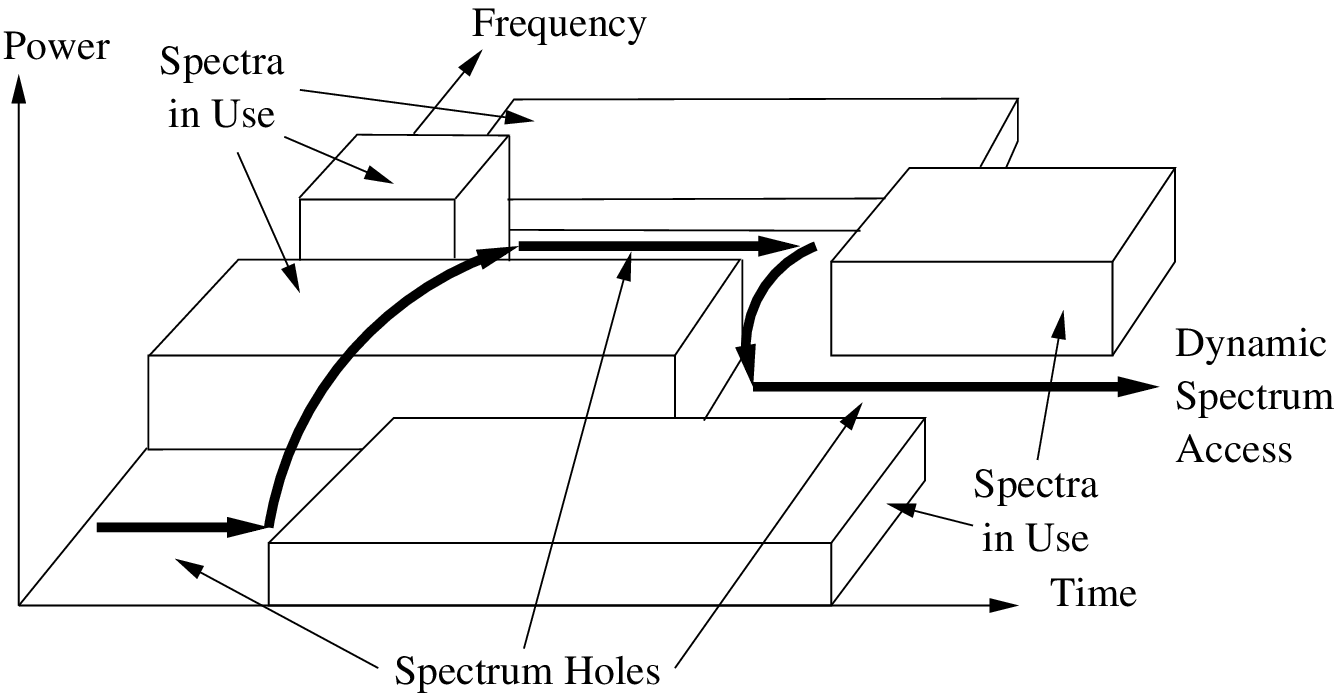}
\vspace{-6mm}
\caption{Holes in frequency and time domains \cite{ALVM06}.}
\label{fig:Holes}
\vspace{2mm}
\includegraphics[width=0.5\textwidth]{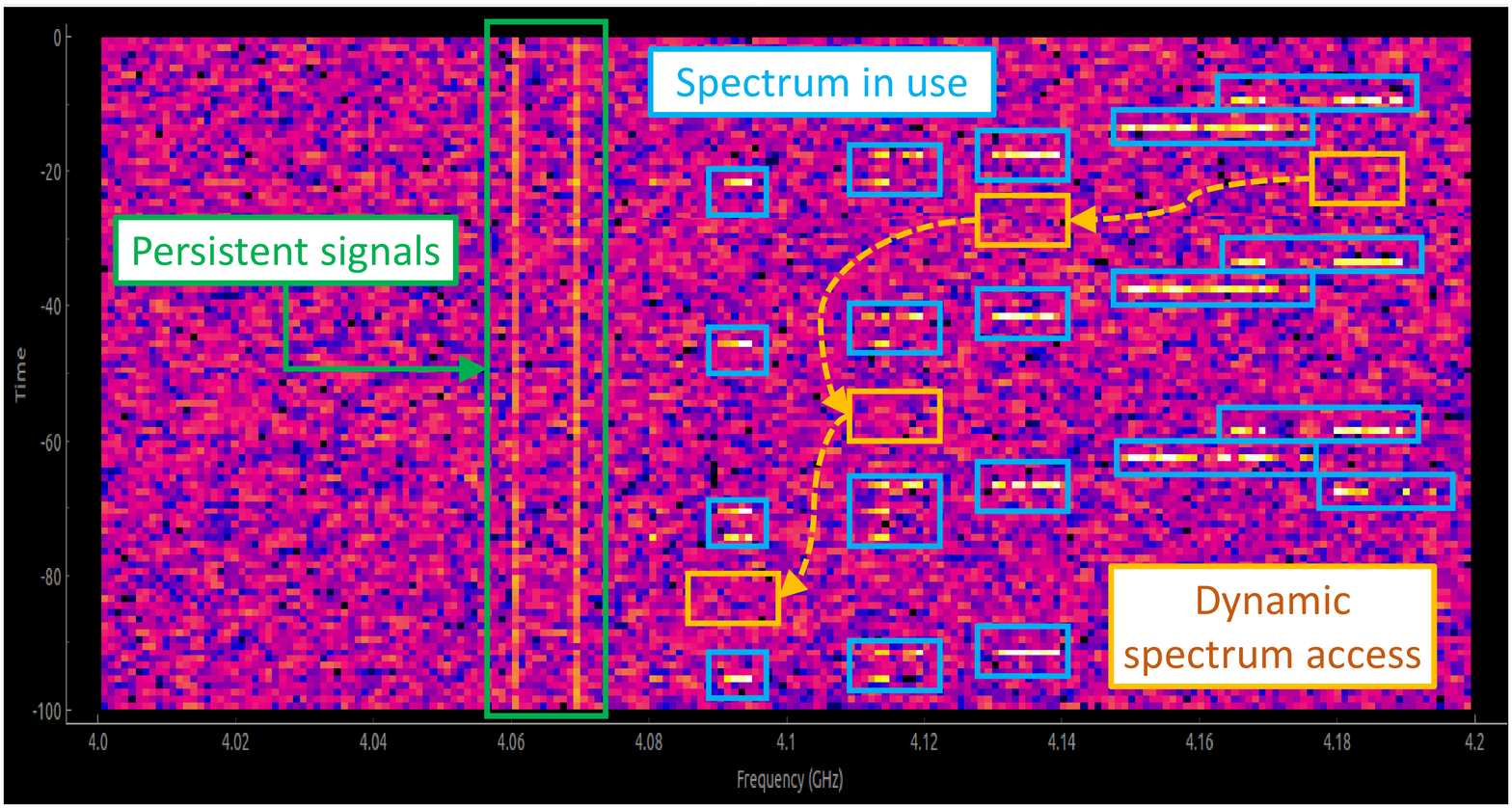}
\caption{Spectrum usage in a shared band and spectrum holes in the frequency (vertical) and time (horizontal) domains enabling dynamic spectrum access. For the figure, spectrum sensing is done by HackRF Software Defined Radio (SDR), and QSpectrumAnalyzer library is used for the real-time data visualization.}
\vspace{-2mm}
\label{fig:Holes2}
\ifCLASSOPTIONonecolumn
\end{wrapfigure}
\else
\end{figure}
\fi
any power, but to offset the interference it causes to the PUs, it relays their transmitted messages. In the interweave technique, the goal is to opportunistically communicate
in the spectrum holes, or more generally the space-time-frequency voids that are not in use by either licensed and unlicensed users. As depicted in Fig.~\ref{fig:Holes},
these voids change with frequency and time. The interweave technique requires periodic monitoring of the spectrum for detection of user activity, so that the SU can transmit opportunistically over the space-time-frequency voids with minimal interference. All of the above approaches can benefit from SUs employing directional antennas or beamforming for more flexible control of the interference. Fig.~\ref{fig:Holes2} depicts spectrum sensing by a Software Defined Radio (SDR) in an actual setting.

For this research topic, implementing the spectrum sensing task using a GAN-based technique as in \cite{DS18} is an interesting and potentially useful first step. It provides {\em domain adaptation\/} by creating synthetic data that enables the classifier to quickly adapt to changes in the spectrum environment.
The approach in \cite{DS18} can be extended to exploit voids in all possible dimensions including space, time, and frequency, which is of particular importance to interweave cognitive radios. To extend the research in this direction, an approach based on CGANs that emphasizes specific subsets of the space, time, and frequency spectrum using specific domain knowledge can be employed. For example, an approach based on CGANs, conditioned on specific subsets of the space, time, and frequency using specific domain knowledge can be employed. Such a CGAN will be able to generate synthetic data that is specific to certain ``labels'' that correspond for example to known locations (e.g., hot spots), time periods (e.g., rush hour), or occupancy patterns of certain frequency bands. A further step in this direction is the investigation of the performance of the algorithms in the presence of low SNR as in \cite{TZL10,VPV12,HSZBWLG17,YHWL19} and parameter uncertainties as in \cite{CS19,YPLRGMC19}.

The approach in \cite{DS18} consists of the following three steps: {\em i)\/} training a GAN using real training samples so that the generator learns to synthesize new samples, {\em ii)\/} training of a classifier with augmented data so that the original and limited training set is expanded with synthetic data, and, {\em iii)\/} regular operation by classifying data using the trained model. It is possible to extend this approach by training a CGAN for more typical scenarios, such as known locations, time periods, and occupancy patterns of certain bands, as discussed in the previous paragraph. In this context, unlike \cite{DS18}, it is possible to make a comparison of this ML-based approach with conventional narrowband spectrum sensing techniques based on energy detection, cyclostationary feature detection, matched filter detection, and wideband spectrum sensing techniques such as compressed sensing \cite{ALB11,AK19} in terms of performance and complexity, specifically in the three areas discussed in Sec.~\ref{sec:introduction}.

In addition to \cite{CHHM21} discussed in Sec.~\ref{sec:augmentation}, another work on the AMC problem is provided in \cite{LLLZ18}. This work uses a modified version of CGAN and AC-GAN, which is operated in a semi-supervised (not unsupervised) mode. The modifications are needed to solve the following problems: {\em i)\/} the complexity of the modulation signals cause non-convergence due to the mapping from the high-dimensional parameter space to the low-dimensional classification space, and {\em ii)\/} lack of diversity of the generated samples cause mode collapse. Mode collapse is defined as the difficulty in convergence, especially during the training process, due to conditional constraints. The modifications consist of an encoder block $E$, a transform block $T$, and the splitting of the classification and discrimination functions of the conventional discriminator $D$ by adding an explicit classifier $C$ \cite[Fig. 1]{LLLZ18}. The encoder $E$ and the transform $T$ are inspired by Conditional Variational Autoencoder (CVAE) \cite{SLY15} and Spatial Transformer Network (STN) \cite{JSZK16}, respectively. The definition of the loss functions involving all the blocks and their optimization expressions are given in \cite[Eq. (10)-(20)]{LLLZ18}. To evaluate the performance of the system, the synthetic dataset in \cite{OW16} is used. This dataset contains signals of eleven modulation modes at SNRs ranging from -20 dB to 20 dB, with channel distortion, frequency offset, phase offset, and Gaussian noise added. Comparisons with a number of well-known deep learning methods show gains in classification accuracy of up to 12\%.

An important implementation technique for NextG networks is Cooperative Spectrum Sensing (CSS), where a number of SUs cooperate to improve sensing performance.
CSS can be implemented using either a centralized or a distributed approach. In centralized sensing, the SUs send their sensing information to a Fusion Center (FC) that makes the sensing decision and broadcasts this information back to all SUs. In distributed sensing, the SUs share sensing information among each other and make independent sensing decisions locally.
In both approaches, the SUs can either transmit their individual sensing decisions or soft information in the form of partial statistics. The information combining schemes used by the FC or SUs can be categorized as either hard or soft combining schemes. The most common hard combining algorithms are AND combining, OR combining, $M$-out-of-$N$ combining, and quantized hard combining, while soft combining algorithms include selection combining, maximal ratio combining, equal gain combining, and square law combining \cite{PP17}. The soft combining approach provides the best detection performance, but at the cost of additional control channel overhead.

It is desirable to extend the technique described above to include CSS. In one implementation, GANs can be placed at the SUs and classifications can be combined at the FC.
An alternative is to have the SUs transmit partial statistics to be used by a single GAN at the FC. In this regard, it is worthwhile to investigate schemes that will employ combinations of centralized versus distributed CSS with hard or soft decisions. The number of combinations is high but it should be possible to reach quick conclusions. The main interest in this activity is to determine the optimal placement for GANs in interweave networks, and how much performance improvement they can provide.

There is an important lesson that can be drawn from this subsection. Due to the tremendous number of users and applications in NextG networks, spectrum sharing will need to be carried out in a dynamic environment employing principles of cognitive networking. The goal is to determine the voids based on time, frequency, and space and exploit them to accomodate more users. Two specific techniques for spectrum sharing based on GANs are discussed in \cite{DS18,LLLZ18} and summarized in this subsection.
\subsection{Detecting Anomalies}\label{sec:detanom}
In a GAN, the generator captures the distribution of the training data, and the discriminator can detect false from real, making a GAN an attractive ML technique for anomaly detection. This observation has led to a number of anomaly detection techniques employing GANs \cite{YGBZTD18,WZL18,LLTCRE18,HSF18,AAB19,DVRMK19,NWLPAL19,IKS19,ZFLMC19,ZLHCY19}. These applications are typically for various forms of image processing. There are some recent applications of using adversarial networks for anomaly detection in cyber-physical systems \cite{LCGN19,LCSJGN19}, fraud detection in banking \cite{ZYWLL18,ZZSXC18,FDPZP19,SSSRB19}, driver assistance systems \cite{QMB19}, air surveillance \cite{Gokgoz}, prognostics and health management in the aeronautics industry \cite{DHS19}, and network traffic anomaly detection \cite{PBRPG20}.

We will briefly discuss \cite{PBRPG20} since it is related to network anomaly detection. This work proposes BiGAN in conjunction with Principal Component Analysis (PCA) for network anomaly detection using the KDDCUP-99 dataset \cite{KDDCUP-99}. This database includes a wide variety of intrusions simulated in a military network environment used for a competition during a conference to build a network intrusion detector, a predictive model capable of distinguishing between ``bad'' connections, called intrusions or attacks, and ``good'' normal connections \cite{KDDCUP-99}.

\ifCLASSOPTIONonecolumn
\begin{table}[!t]
\centering
\small
\begin{tabular}{lccc}
{\bf Model}&{\bf Precision}&{\bf Recall}&{\bf F1 Score}\\
\hline
\hline
OC-SVM \cite{ZCLZ16}&0.7457&0.8523&0.7954\\
\hline
DSEBM-r \cite{ZCLZ16}&0.8521&0.6472&0.7328\\
\hline
DSEBM-e \cite{ZCLZ16}&0.8619&0.6446&0.7399\\
\hline
DAGMM-NVI \cite{ZSMCLCC18}&0.9290&0.9447&0.9368\\
\hline
BiGAN \cite{ZFLMC18}&0.9363&0.9512&0.9437\\
\hline
PCA+BiGAN \cite{PBRPG20}&0.9442&0.9592&0.9516\\
\end{tabular}
\caption{Performance of the five anomaly detection algorithms on the KDDCUP-99 dataset \cite{PBRPG20}. OC-SVM: One-Class Support Vector Machine \cite{ZCLZ16}, DSEBM-r: Deep Structured Energy-Based Model - using reconstruction error \cite{ZCLZ16}, DSEBM-e: Deep Structured Energy-Based Model - using energy \cite{ZCLZ16}, DAGMM-NVI: Deep Autoencoding Gaussian Mixture Model with Neural Variational Inference \cite{ZSMCLCC18}.}
\label{tbl:PBRPG20}
\end{table}
\else
\begin{table}[!t]
\centering
\small
\vspace{3mm}
\begin{tabular}{lccc}
{\bf Model}&{\bf Precision}&{\bf Recall}&{\bf F1 Score}\\
\hline
\hline
OC-SVM \cite{ZCLZ16}&0.7457&0.8523&0.7954\\
\hline
DSEBM-r \cite{ZCLZ16}&0.8521&0.6472&0.7328\\
\hline
DSEBM-e \cite{ZCLZ16}&0.8619&0.6446&0.7399\\
\hline
DAGMM-NVI \cite{ZSMCLCC18}&0.9290&0.9447&0.9368\\
\hline
BiGAN \cite{ZFLMC18}&0.9363&0.9512&0.9437\\
\hline
PCA+BiGAN \cite{PBRPG20}&0.9442&0.9592&0.9516\\
\end{tabular}
\caption{Performance of the five anomaly detection algorithms on the KDDCUP-99 dataset \cite{PBRPG20}. OC-SVM: One-Class Support Vector Machine \cite{ZCLZ16}, DSEBM-r: Deep Structured Energy-Based Model - using reconstruction error \cite{ZCLZ16}, DSEBM-e: Deep Structured Energy-Based Model - using energy \cite{ZCLZ16}, DAGMM-NVI: Deep Autoencoding Gaussian Mixture Model with Neural Variational Inference \cite{ZSMCLCC18}.}
\label{tbl:PBRPG20}
\end{table}
\fi

A number of works exist that study anomaly detection in cognitive radio networks with conventional methods \cite{ASH10,BRRA13,HYL15,YZQY18,WG18,Dong18,KJKKK19} and with ML \cite{RMLP18,TJMR18,LXWZZ19,RMLSP19,RLMP20,TKFQMGR20,TKMGR20}. We refer the reader to \cite{TKMGR20} for brief descriptions of \cite{TJMR18,LXWZZ19,RMLSP19,RLMP20,TKFQMGR20}. These works provide an introduction to this field, with the important observation that GAN-based methods tend to provide better results in general. However, it is certain that existing research in this area is still limited, and this provides an opportunity for significant contributions.

In \cite{RMLP18}, an adversarial autoencoder was implemented for wireless spectrum anomaly detection \cite{MSJGF16}, which uses ideas similar to a GAN. We will go beyond this approach and leverage the advantages provided by AnoGAN, EGBAD, and GANomaly. We provide details here on an approach using GANomaly, although similar anomaly scores can be defined for other architectures. Consider a collection of received data vectors $X_N$, whose elements are for example the I and Q components of a digitally modulated signal, or vectors of the sampled power spectral density. We seek a model that learns the source distribution $p(X_N)$, so that we can detect when the vector's distribution is different from $p(X_N)$. This is a hypothesis testing problem, where for each vector ${\bf x}\in X_T$ in the test dataset $X_T$, the two hypotheses are
\begin{equation}
\begin{aligned}
H_N\mkern-6mu : &\ \textrm{Sample\ data\ comes\ from\ }p(X_N),\\
H_A\mkern-6mu : &\ \textrm{Sample\ data\ does\ not\ come\ from\ }p(X_N).
\end{aligned}
\end{equation}
We have three assumptions: First, the probability of anomalous behavior in dataset $X_N$ is very low; second, no explicit anomaly labeling is done on the test dataset; and third, no feature extraction is performed before feeding the data to the model. The key insight in employing GAN-based anomaly detection is to bring the data to the latent space, which captures relevant features that can be used to reconstruct the actual input data, with a small anomaly score for normal data. A GAN is trained using (\ref{eqn:V(D,G)}) and (\ref{eqn:minmax}), and during the test and operation stage, (\ref{eqn:Ladv})--(\ref{eqn:anomalyscore}) are used. Once the training process is complete, the model weights are frozen and new data are input into the model. Anomalies are detected based on the anomaly score or the reconstruction loss of the model.

The term anomaly covers a very broad range of possibilities in the transmitted signal. Since the field of ML progresses based primarily on experimentation, a number of datasets should be created to study the performance of the algorithms under consideration. These can be in the form of time series corresponding to I/Q data from a radio receiver after downshifting in frequency and mixing with the local oscillator. This signal includes the effects of fading and multipath. To create anomalous test signals, there are a wide variety of possibilities. Reference \cite{RMLP18} considered four types of normal signals and four types of anomalies. The normal signals were generated as either {\em i)\/} a single continuous signal with random bandwidth, SNR, and center frequency, {\em ii)\/} pulsed signals with parameters similar to {\em i)\/}, {\em iii)\/} multiple continuous signals with possible frequency overlap, or {\em iv)\/} signals with random bandwidths and SNR with deterministic shifts/hops in frequency. The anomalous signals were chosen as either {\em i)\/} the same as normal signals in {\em i)\/} above, {\em ii)\/} random pulsed transmissions in the given band, {\em iii)\/} pulsed wideband signals covering the entire frequency band, or {\em iv)\/} signals from other classes in the synthetic dataset. Normal and anomalous signals such as these can be considered, as well as other random signal types with arbitrary power spectral densities.

Anomaly detection within the context of cognitive radio in the mmWave band is studied in \cite{TKMGR20}, where the authors generate their own datasets with a number of anomalies and evaluate three methods including two based on GANs for their detection. As mmWave radios are vulnerable to malicious users due to the shared access medium and the complex radio environment, this opens up new threat possibilities. Moreover, since several security critical applications in the NextG networks such as the Vehicle-to-Everything (V2X) networks are based on mmWave communications, this problem is particularly important. In addressing this problem, \cite{TKMGR20}
makes use of three ML techniques: Variational Autoencoder (VAE) \cite{KW19}, CGAN \cite{MO14}, and AC-GAN \cite{OOS16}. This work created its own dataset by using mmWave equipment in a laboratory environment with implementation using Field Programmable Gate Arrays (FPGAs) and building blocks such as local oscillators, intermediate frequency modules, mmWave radio heads, horn antennas, etc. The operation is at 28 GHz by using Cyclic-Prefix Orthogonal Frequency Division Multiplexing (CP-OFDM) with
BPSK, QPSK, 16-QAM, and 64-QAM. Complex I/Q data is collected at baseband after the downconversion process. There are eight channels with 100 MHz bandwidth. The normal behavior is a fixed signal which occupies channel 4 and a sequentially moving signal among channels 8, 6, 3, and 1. This normal behavior is used in the training phase. For testing, three modalities are introduced. Modality 1 is a fixed signal in channel 4, and a moving signal that jumps among channels 5, 7, 2, and 5. Modality 2 is a fixed signal in channel 4 and a moving signal that jumps among channels 7, 5, 2, and 1. Modality 3 is a fixed signal in channel 4 and a moving channel that jumps among channels 5, 7, 6, and 5. %In the experiment, $P_D$ and $P_{FA}$ are calculated and ROCs ($P_D$ vs. $P_{FA}$) are drawn.
Performance results are provided by AUROC and ACC.
These results are given in Table~\ref{tbl:TKMGR20-1}.
\ifCLASSOPTIONonecolumn
\begin{wraptable}{r}{0.5\textwidth}
\centering
\small
\setstretch{1.0}
\vspace{-3mm}
\begin{tabular}{lc|cc}
&{\bf Modality}&{\bf AUROC}&{\bf ACC}\\
\hline
\hline
&{1}&0.9365&0.9356\\
{\bf VAE}&{2}&0.9577&0.9551\\
&{3}&0.9232&0.9382\\
\hline
&{1}&0.9566&0.9657\\
{\bf CGAN}&{2}&0.9737&0.9696\\
&{3}&0.9545&0.9668\\
\hline
&{1}&0.9741&0.9804\\
{\bf AC-GAN}&{2}&0.9751&0.9757\\
&{3}&0.9742&0.9660\\
\end{tabular}
\caption{AUROC and ACC performance of the three ML algorithms.}
\label{tbl:TKMGR20-1}
\vspace{2mm}
\begin{tabular}{l|c|c}
&{\bf Training} [mm:ss]&{\bf Testing} [mm:ss]\\
\hline
{\bf VAE}&15:09&01:00\\
{\bf CGAN}&15:16&01:36\\
{\bf AC-GAN}&30:42&03:16\\
\end{tabular}
\caption{Computational times of the three ML algorithms.}
\vspace{-1mm}
\label{tbl:TKMGR20-2}
\end{wraptable}
\else
\begin{table}[!t]
\centering
\small
\vspace{3mm}
\begin{tabular}{lc|cc}
&{\bf Modality}&{\bf AUROC}&{\bf ACC}\\
\hline
\hline
&{1}&0.9365&0.9356\\
{\bf VAE}&{2}&0.9577&0.9551\\
&{3}&0.9232&0.9382\\
\hline
&{1}&0.9566&0.9657\\
{\bf CGAN}&{2}&0.9737&0.9696\\
&{3}&0.9545&0.9668\\
\hline
&{1}&0.9741&0.9804\\
{\bf AC-GAN}&{2}&0.9751&0.9757\\
&{3}&0.9742&0.9660\\
\end{tabular}
\caption{AUROC and ACC performance of the three ML algorithms.}
\label{tbl:TKMGR20-1}
\vspace{2mm}
\begin{tabular}{l|c|c}
&{\bf Training} [mm:ss]&{\bf Testing} [mm:ss]\\
\hline
{\bf VAE}&15:09&01:00\\
{\bf CGAN}&15:16&01:36\\
{\bf AC-GAN}&30:42&03:16\\
\end{tabular}
\caption{Computational times of the three ML algorithms.}
\vspace{-1mm}
\label{tbl:TKMGR20-2}
\end{table}
\fi
CGAN provides better performance than VAE and AC-GAN provides better performance than CGAN, namely, VAE performs an AUROC between 0.9365 and 0.9577, CGAN between 0.9545 and 0.9737, and AC-GAN between 0.9741 and 0.9751. Reference \cite{TKMGR20} attributes the better performance by GANs to the fact that they learn the relation between random noise and generated data such that
the generated data is close to real data, and as a result, they can capture the dynamics in the real data. Whereas, VAEs return the posterior probability that an observation belongs to a cluster. They do so by learning the latent vector corresponding to an input, and as a result, all the dynamics in the real data that VAEs have access to are limited to the latent vectors already modeled. Due to a larger degrees of freedom with the former, it is generally considered that GANs can have better performance. Computational times for the three algorithms, performed on an NVIDIA GeForce GTX 1080 Ti GPU, are given in Table~\ref{tbl:TKMGR20-2}. Clearly, there is a big price paid for the better performance of AC-GAN (15 minutes vs. 30 minutes).

In \cite{SDSHFG19}, a deep learning based signal or modulation classification solution is described, where {\em i)\/} signal types may change over time, {\em ii)\/} some signal types are not known {\em a priori,\/} and therefore there is no training data available, {\em iii)\/} signals are potentially spoofed such as smart jammers replaying other signal types, {\em iv)\/} different signals may be superimposed due to interference from concurrent transmissions. The authors present a CNN that classifies the received I/Q samples as idle, in-network signal, jammer signal, or out-network signal. Traditional approaches for signal classification require expert design or knowledge of the signal. Modulations are classified into {\em i)\/} idle, {\em ii)\/} in-network user signal, {\em iii)\/} jamming signals, and {\em iv)\/} out-network user signals where there are 3, 4, and 3 different modulation techniques corresponding to {\em ii)\/}--{\em iv)\/}. There are in-network users who try to access the channel opportunistically (SUs), out-network users with priority channel access (PUs), and jammers that all coexist. The authors use the dataset in \cite{OCC16}. There are ten modulations with SNRs from -20 dB to 18 dB in 2 dB increments. Radio fingerprinting via radio hardware imperfections such as I/Q imbalance, time or frequency drift, and power amplifier effects are used to identify the type of a transmitter. In addition to a CNN structure, Minimum Covariance Determinant (MCD) , $k$-means clustering, and Independent Component Analysis (ICA) techniques are employed. Results demonstrate the feasibility of using deep learning to classify RF signals with high accuracy in unknown and dynamic spectrum environments. By using the signal classification results a distributed scheduling protocol is developed where SUs share the spectrum with each other while avoiding interference imposed to PUs and received from jammers.

The use of GANs that exploit the spatial domain to identify anomalous signal sources can be considered, using data received from antennas in different locations. In many applications, normal network operations are consistent with users transmitting from specific locations, such as in a stadium setting, a large hall, a shopping mall, etc. A GAN trained in this setting with an array of antennas or widely separated receivers can be used to differentiate between normal network traffic and signals that arrive from anomalous directions, even if their spectral characteristics are identical to normal users. A similar approach can be used to detect differences between mobile and stationary sources, even if no detectable Doppler shift can be measured. Mobile sources typically have lower power, reduced persistence, a time-varying polarization, as well as time-varying ranges and azimuth/elevation angles. Further differentiation is possible between ground-based and airborne sources. A particularly important direction is to investigate whether incorporating the spatial dimension together with the time and frequency dimensions can significantly improve the ability of GANs to detect anomalous network behavior.

In passing, we would like to state that more research is needed to understand the performance vs. computational complexity of autoencoders for anomaly detection \cite{BWAN18,BPB19,VUHMPS19,NLDLC19,GK19}, especially in terms of their performance for given computational complexity. Tables~\ref{tbl:TKMGR20-1} and \ref{tbl:TKMGR20-2} provide an interesting comparison in this regard.

The most important lesson learned from this subsection is that the use of GANs result in powerful techniques for anomaly detection in wireless networks. In Sec.~\ref{sec:initres}, we will expand this observation and provide more evidence to this fact by adopting a GAN-based anomaly detection algorithm for NextG wireless applications.
%
%\subsection{Mitigating Security Attacks}\label{sec:security}
%
\subsection{Security Applications of GANs in NextG}\label{sec:security}
Another important application area of GANs is wireless security. In these applications, ML can be used to create an attack, defend against an attack, or both. Some specific cases are discussed below.

In \cite{RMCP19}, a laboratory study is carried out with eight Universal Software Radio Peripheral (USRP) radios as trusted transmitters. The goal is to use the imbalance in the detected I/Q components due to unique hardware differences in each transmitter for fingerprinting, i.e., identifying which transmitter is which. The authors use the generator $G$ of a conventional GAN to create fake transmitters, and then the GAN classifier attempts to distinguish between real and fake transmitters, achieving an accuracy of 99.9\%. The authors also develop approaches to distinguish between the legitimate transmitters, one based on a CNN that obtains an accuracy of 81.6\%, and another based on a Deep Neural Network (DNN) that scores 96.6\%. Although this is a simple experiment, it shows the power of ML in wireless security.

A wireless spoofing algorithm based on a GAN architecture is proposed in \cite{SDS19}, split between an adversary transmitter $A_T$ and an an adversary receiver $A_R$ placed close to the actual receiver $R$. Feedback from $A_R$ based on its location allows $A_T$ to know the channel between the actual transmitter $T$ and $R$ as well as between $A_T$ and $R$ with sufficiently high accuracy. Then, a GAN is implemented between $A_T$ and $A_R$ in which the generator $G$ is trained at $A_T$ and the discriminator $D$ is implemented at $A_R$. $A_T$ in turn adjusts its transmission parameters such that its signal will appear to $R$ as if they come from $T$. In this process, $A_R$ trains the discriminator $D$ such that the classification error is minimized, i.e., it attempts to achieve
\begin{equation}
\min_D \mathbb{E}_{{\bf z}\sim p_{\bf Z}({\bf z})}[\log(1-D(G({\bf z})))]-\mathbb{E}_{{\bf x}\sim p_{\rm data}}[\log(D({\bf x}))].
\end{equation}
At the same time, $A_T$ trains the generator $G$ such that the classification error is maximized, i.e., it attempts to achieve
\begin{equation}
\max_G \mathbb{E}_{{\bf z}\sim p_{\bf Z}({\bf z})}[\log(1-D(G({\bf z})))]. %-\mathbb{E}_{{\bf x}\sim p_{\rm data}}[\log(D({\bf x}))]. % KD: removed this part
\label{eqn:A_RG}
\end{equation}
The process is continued until convergence. Thus, although they are at different locations, $A_T$ and $A_R$ train $G$ and $D$ while playing the following minimax game
\ifCLASSOPTIONonecolumn
\begin{equation}
\min_D\max_G \mathbb{E}_{{\bf z}\sim p_{\bf Z}({\bf z})}[\log(1-D(G({\bf z})))]-\mathbb{E}_{{\bf x}\sim p_{\rm data}}[\log(D({\bf x}))].
\end{equation}
\else
\begin{equation}
\begin{split}
\min_D\max_G & \mathbb{E}_{{\bf z}\sim p_{\bf Z}({\bf z})}[\log(1-D(G({\bf z})))]\\
             & -\mathbb{E}_{{\bf x}\sim p_{\rm data}}[\log(D({\bf x}))].
\end{split}
\end{equation}
\fi
Simulation results show that, while the probability of successful spoofing is only 7.9\% for random signals and 36.2\% in an amplify-and-forward architecture, the GAN is able to achieve a success rate of 76.2\%.

The problem of thwarting an Intrusion Detection System (IDS) is studied in \cite{UALQA19}. While this work is not specifically about wireless networks, the basic results are still applicable to wireless systems. While many ML approaches have been proposed for IDS, the authors of \cite{UALQA19} show that such systems are vulnerable to an attack employing a GAN. On the other hand, the paper also shows that if the IDS system based on conventional ML is replaced by one based on GANs, it can be made more robust against adversarial perturbations.

In \cite{YL19}, an interesting problem involving two Unmanned Aerial Vehicles (UAVs) trying to communicate in the presence of an active jammer that eavesdrops their transmissions and jams only when the two are at communicating in the same frequency band is described. A three-way GAN is developed with the generator $G$ present at the jammer and two classifiers $D_s$ and $D_r$ present, one at each UAV. The generator $G$ operates without access to the classifiers. For reasons of stability, \cite{YL19} does not employ the conventional $\log$-based formulation for driving the backpropagation algorithm, but instead, another version based on a least squares formulation called LSGAN \cite{MLXLWS17}. The performance of the algorithm is compared against two GAN-based algorithms with three players and one non-GAN-based game theoretic approach. The algorithm in \cite{YL19} is shown to outperform the other three in terms of average connection latency, attack probability, and packet delivery ratio in the presence of channel switching and jamming. This paper shows the sophistication of GAN-based approaches in wireless applications.

In addition to the GAN-based wireless security work described above, there are many other papers in the literature that cover various aspects of wireless security by employing other ML techniques, in terms of both designing attacks and ways to mitigate them. A sample listing is \cite{ALNT14,YBHKSV17,WKSS17,DGZKM18,SSEDLL18,CDMS18,KG18,YRZZ18,SL18,ZSF18,LHGNWLK18,RAAKNC18,AFP18,CFCS18,ESS18,SESL19,SL19,HGG19,FBH19,SSE19,HDL19,FWT19,ZRMP19,SDS21}.

For spectrum sharing applications, not only traditional security threats such as receiver jamming, but also newly emerged cognitive-radio-specific security threats such as PU Emulation Attacks (PUEA), SSDF, common control channel jamming, selfish users, and intruding nodes should be considered. A detailed description of these threats can be found in \cite{CG08,ZXG08,Esch12,ATVYL12,HR16}.

Cognitive radio networks rely on a trustworthy spectrum sensing process, the key to which is the ability to distinguish PU signals from SU signals in a robust way. The use of licensed spectrum bands by PUs may be sporadic, so an SU must constantly monitor for the presence of PUs in candidate bands. If an SU detects the presence of a PU in the band under observation, it cannot use the current band. If there is no PU currently active in the band, then SUs employ a medium access control mechanism to share the band. In general, it is expected that no change will be required on the part of the PU transmission system or air interface in order to accommodate the SUs, e.g., \cite{FCC03}, and thus it is the responsibility of the SUs to correctly identify PUs with their existing air interface. As a result, an SU's ability to correctly identify PUs becomes very important not only for avoiding interference to the PUs, but also to be able to increase their own throughput. Distinguishing whether or not a user is a PU is nontrivial, especially when rogue users modify their air interface in order to mimic a PU's signal characteristics. This is known as a PUEA \cite{CP06}.

In order to mitigate the PUEA problem, it is possible to draw parallels with a phenomenon known for adversarial attacks against classifiers that employ neural networks. This phenomenon became apparent in classification problems in conventional fields such as image processing, where it was observed that certain minor variations in the input, undetectable by humans, can make a classifier fail, see e.g., \cite{SZSBEGF14,GSS15}. An example that appeared in \cite{GSS15} is very well-known. Using a 22-layer deep CNN that is able to correctly identify a panda with 57.7\% accuracy, adding a slight amount of noise results in it being characterized as a gibbon with 99.3\% confidence. This vulnerability of deep neural networks to adversarial attacks has been well-documented, see e.g., \cite{PMJFCS15,KGB17,CW17,CSZYH18,MMSTV19}.

A number of approaches exist to deal with the problem of adversarial attacks in the context of image processing; a good review is available in \cite{QLZW19}. There is a recent study which finds that generative classifiers, such as GANs, are actually more robust to adversarial attacks \cite{LBS19}. Given the success of ML techniques developed for image processing and successfully adapted to problems in wireless networking, it is possible to leverage some of this prior work in the context of designing GAN-based ML systems to counteract adversarial attacks. The first such technique is described in \cite{SKC18}, and is referred to as Defense-GAN. In this technique, the GAN is trained to model the distribution of unperturbed data. At inference time, it finds an output that does not contain any adversarial changes that is ``close'' in some sense to the given input data. This output is then fed to the classifier, which can employ any model, and the result is used as the classification of the data. This technique does not assume knowledge of the process for generating the adversarial examples. Reference \cite{SKC18} empirically shows that Defense-GAN is consistently effective against different attack methods and improves on existing defense strategies.

\ifCLASSOPTIONonecolumn
\begin{figure}[!t]
\else
\begin{figure*}[!t]
\fi
%\begin{wrapfigure}{r}{0.65\textwidth}
\vspace{-1mm}
\centering
\ifCLASSOPTIONonecolumn
\scalebox{0.7}{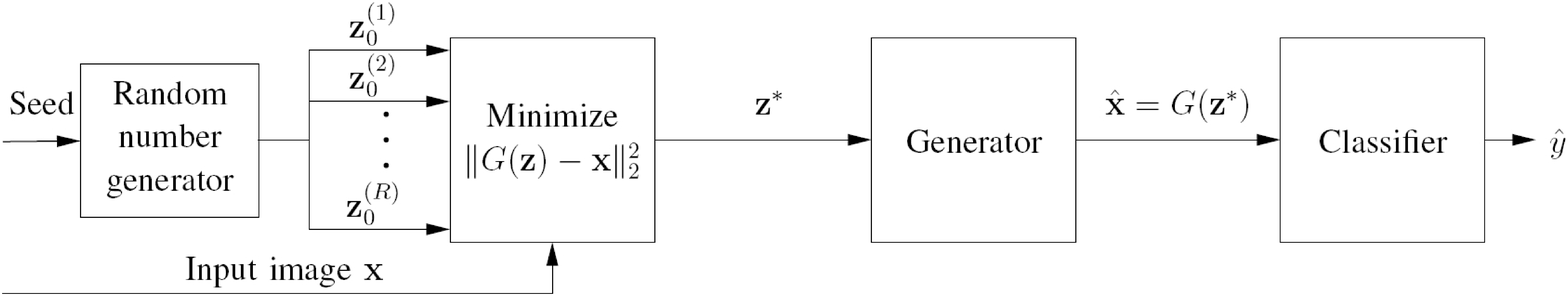}
%\vspace{3mm}
\else
\includegraphics[width=0.65\textwidth]{Figures/Defense-GAN.eps}
\fi
\caption{Defense-GAN algorithm \cite{SKC18}.}
\vspace{-1mm}
\label{fig:DefenseGAN}
%\end{wrapfigure}
\ifCLASSOPTIONonecolumn
\end{figure}
\else
\end{figure*}
\fi
Fig.~\ref{fig:DefenseGAN} illustrates the basic idea in \cite{SKC18}. For a given input observation {\bf x}, the system finds the closest element (in the $L_2$ norm) of the data space that can be produced by the generator operating on an element of the latent space. The resulting element of the data space, presumably free from the affects of the adversarial perturbation, is then classified in the regular way. Intuitively, if ${\bf x}$ is not perturbed, then the GAN can generate a sufficiently close local copy of it to be classified as ``real.'' On the other hand, a perturbed ${\bf x}$ will not be among the possible outputs of the generator $G$ trained by using unperturbed data, and the classifier will declare it as ``fake.'' There is a mathematical development that justifies this intuition in \cite{SKC18}. The paper shows that the Defense-GAN technique is very effective against both black-box and white-box attacks, i.e., attacks that do not have any knowledge of $G$ and the classifier $C$, as well as those that do have full knowledge of $G$ and $C$, respectively. Other works that use a similar concept based on GANs are available in \cite{SG18,LKK19}.

In \cite{SDS21}, the effects of a GAN-based spoofing attack to generate synthetic wireless signals that cannot be statistically distinguished from intended transmissions is studied. The spoofing attack can be used for various adversarial purposes such as emulating PUs in cognitive radio networks and fooling signal authentication systems to intrude protected wireless networks. The adversary is modeled as a pair of transmitter and a receiver that build the generator and discriminator of a GAN, respectively. The adversary transmitter trains a deep neural network to generate the best spoofing signals and fool the best defense trained as another deep neural network at the adversary receiver. Thus, the generator and the discriminator of the GAN are at different locations, collaborating over the air such that the GAN-generated signals cannot be reliably discriminated from intended signals. The adversary and the defender may have multiple transmitter or receiver antennas. Spoofing is accomplished by jointly capturing waveform, channel, and radio hardware effects inherent to wireless signals under attack. It is demonstrated in \cite{DS18} that a GAN-based spoofing attack can be successfully performed and its performance is compared with random signal attack and replay attack. An important observation is that as the attacker's transmitter gets close to the transmitter, the likelihood of the attacker to generate high-fidelity synthetic signals for the spoofing attack increases. In summary, the sussess probability of the GAN-based spoofing attack is very high. This holds for different network topologies and when node locations change from training to test time. This probability improves further when multiple antennas are used at the transmitter.

It is possible to devise defensive strategies against such attacks. A simple strategy is to change the idle and busy labels, so that the legitimate transmitter may not transmit if it senses the channel as idle, or it may transmit even if the channel is sensed as busy \cite{SSEDLL18,ESS18}, assuming that the unobserved transmitter is not a PU. This can be done over a subset of the transmit opportunities. For example, by introducing 10\% of false labels, the legitimate transmitter, as the defender, can increase the misdetection performance of the attacker from 3.95\% to 20.79\% and the false alarm rate from 18.10\% to 25.16\%. The results for a single-input single-output (SISO) scenario from \cite{SSEDLL18} are shown in Table~\ref{tbl:SSEDLL18-2}. Clearly, this simple algorithm can result in a significant improvement.

\ifCLASSOPTIONonecolumn
\begin{table}[!t]
\else
\begin{table*}[!t]
\fi
\vspace{1mm}
\centering
\small
\begin{tabular}{c|c|c|c|c}
{\bf \# of defense operations} & \multicolumn{2}{c|} {\bf Attacker error probabilities} & \multicolumn{2}{c} {\bf Transmitter performance}\\
\cline{2-5}
{\bf / \# of all samples} & {\bf Misdetection} & {\bf False alarm} & {\bf Throughput} & {\bf Success ratio} \\
\hline
\hline
0\% (no defense) & 3.95\% & 18.10\% & 0.012 & 2.91\% \\
\hline
10\% & 20.79\% & 25.16\% & 0.074 & 17.13\% \\
\hline
20\% & 33.88\% & 40.69\% & 0.124 & 28.84\% \\
\hline
30\% & 40.09\% & 44.79\% & 0.170 & 35.42\% \\
\hline
40\% & 45.18\% & 43.75\% & 0.206 & 41.53\% \\
\hline
50\% & 41.63\% & 45.10\% & 0.204 & 39.23\% \\
\end{tabular}
\caption{Results for defensive strategies against jamming attacks based on adversarial learning \cite{SSEDLL18}.}
\label{tbl:SSEDLL18-2}
\ifCLASSOPTIONonecolumn
\end{table}
\else
\end{table*}
\fi
It is possible to extend this work to the case where NACKs, in addition to ACKs, are employed. ACKs are only positive acknowledgements, they do not reveal if there was a transmission. NACKs state there was a transmission but it was not received correctly, possibly due to a collision, but also potentially due to other effects such as fading. This increases the state space of the learning algorithm and may improve the performance of the attacker. A potential direction is to investigate if there is a simple algorithm similar to the one above that provides a defense against such attacks.

For the case of cooperative spectrum sensing, a recent study makes adversarial use of ML to construct a surrogate of the fusion center's decision model \cite{LZLXS19}. The authors then propose an algorithm to create malicious sensing data, and show that this type of attack is very effective. Reference \cite{LZLXS19} shows, {\em via experiments,\/} that with existing defenses it can achieve up to an 82\% success ratio while only manipulating a small number of malicious nodes. Then, a mechanism referred to as an influence-limiting policy is introduced which achieves a disruption ratio reduction of up to 80\% of the attacks introduced in \cite{LZLXS19}. An important observation that can be made based on this work is that, as in many security problems, specific types of attacks require specific types of solutions. It is possible to pursue attacks similar to those in \cite{LZLXS19} and attempt to develop relevant solutions. However, it should be emphasized that a key aspect of any defense against a particular attack is first detecting that an attack has occurred, which is exactly the anomaly detection problem discussed above. Strong anomaly detection is the first step in mitigating attacks in spectrum sensing problems, both cooperative and noncooperative. To that end, it is worthwhile to investigate the effectiveness of the GAN-based anomaly detection techniques discussed in Sec.~\ref{sec:anomalydetection} against a number of different attacks such as those described above.

While many conventional countermeasures against security attacks in cognitive radio networks exist \cite{BSC13}, there is very limited published work on ML techniques for this purpose \cite{SSM19}. Thus, this is a potentially very fruitful research area.

The most important lesson learned from this subsection is that GANs can be used for security attacks or as a mechanism for defense against any security attacks. We have discussed seven works from the literature towards both ends.

Table~\ref{tbl:GANs-NextG} summarizes all of the works from the literature discussed in Sec.~\ref{sec:proposed}.

\ifCLASSOPTIONonecolumn
\begin{table}[!t]
\else
\begin{table*}[!t]
\fi
\small
\centering
\begin{tabular}{|c|l|l|}
\hline
Reference & Application & Comment \\
\hline
\cite{DS18} & Spectrum Sharing & Spectrum sensing with domain adaptation and data augmentation.\\
\cite{LLLZ18} & & AMC solution via modified CGAN and semi-supervised AC-GAN.\\
\hline
\cite{PBRPG20} & Anomaly Detection & Use of BiGAN and PCA for anomaly detection. Comparisons in Table~3.\\
\cite{RMLP18} & & Unsupervised spectrum anomaly detection via adversarial autoencoder.\\
\cite{TKMGR20} & & For mmWave. Comparison of VAE, CGAN, AC-GAN in Tables 4-5.\\
\cite{SDSHFG19} & & Modulation classification via CNN: Idle, in- or out-network, jammer.\\
\hline
\cite{RMCP19} & Security & Create and then classify fake transmitters via GAN. 99.9\% accurate.\\
\cite{SDS19} & & Spoofing with a GAN where its $G$ and $D$ are not colocated. High success rate.\\
\cite{UALQA19} & & Intrusion detection with improved attack or defense by GANs. \\
\cite{YL19} & & Three-way GAN among a jammer and two UAVs trying to communicate.\\
\cite{SKC18} & & Defense-GAN (Fig. 7). Effective against both black- and white-box attacks.\\
\cite{SDS21} & & Effects of GAN-based spoofing to generate synthetic signals like real ones.\\
\cite{ESS18} & & Strategy based on changing labels against GAN-based spoofing attacks. Table~6.\\
\hline
\end{tabular}
\caption{Different implementations of techniques using GANs for spectrum sharing, anomaly detection, and security applications in NextG. The main characteristic of each application is specified as a comment. For a detailed discussion of each of the techniques,
see the text.  PCA: Principal Component Analysis, CNN: Convolutional Neural Network, UAV: Unmanned Aerial Vehicle.}
\label{tbl:GANs-NextG}
\ifCLASSOPTIONonecolumn
\end{table}
\else
\end{table*}
\fi
\section{Simulation Results}\label{sec:initres}
The problem of outlier detection in signal classification
is a critically challenging one for NextG networks. Signal classification is needed for various NextG applications, including user equipment (UE) identification, spectrum sharing and coexistence (as in the CBRS band), and
jammer (interference) detection.
Specifically, for the case where there is no data about unknown waveforms (that can be considered as outliers), the
signal classification
problem becomes even more challenging. In this case, in general, the outlier detection algorithms are trained using only inlier waveforms and tested with inliers and outliers. Recent studies have shown that GANs can help solve the outlier detection problem. AnoGAN has been proposed as an unsupervised anomaly detection algorithm to calculate an anomaly score \cite{SSWSL17}. Despite its good performance and relaxed assumption that does not require labeled data, it suffers from instabilities in GAN training. In addition, AnoGAN is an iterative algorithm which hinders its real-time application. Fast AnoGAN (f-AnoGAN) extends AnoGAN by using a more stable GAN architecture, WGAN, and it does not require iterations (only one forward pass) \cite{SSWLS19}. F-AnoGAN uses an encoder to obtain the latent features of a generator. The latent features are input to the generator. The anomaly score is calculated as the mean squared error (MSE) of the input sample and its reconstruction
$\|{\bf x} - G(E({\bf z})) \|_2^2$ plus the MSE of the discriminator features of the original and those of the reconstructed signal, $\| D_f({\bf x})-D_f(G(E({\bf x})))\|_2^2$, weighted by a scaling factor $\kappa$.

This approach can be applied to anomaly detection in RF spectrum data. As an initial result, we used the RFMLS 2016a dataset \cite{OCC16} that includes eleven different modulations collected over a wide SNR range between -20 dB and 18 dB. The f-AnoGAN algorithm is trained on only one (known) modulation and evaluated using all eleven modulations. We expect the model to label samples from the known (trained) modulation as inlier and samples from all others as outliers. In contrast to \cite{GAADC17} that uses WGAN, we used the WGAN-GP which is known to be more stable. The f-AnoGAN model outputs an anomaly score per sample and the AUROC figures are calculated.

In our evaluations, the f-AnoGAN model is trained in three steps:
\begin{enumerate}
    \item First, we trained the generator and discriminator of the WGAN-GP model, both networks are summarized in Table~\ref{tab:generator_model} and Table~\ref{tab:discriminator_model} in the Appendix, respectively.
    \item Second, we trained the encoder (architecture presented in Table~\ref{tab:encoder_model} in the Appendix).
    \item Finally, we evaluated the performance of the anomaly detection using the anomaly score.
\end{enumerate}

This three-step procedure closely follows the f-AnoGAN implementation of \cite{SSWLS19}.\footnote{https://github.com/tSchlegl/f-AnoGAN}

In evaluating the fidelity and diversity of the generative models, we compared the use of five different measures \cite{NOUCY20}:
{\em i)\/}	sum of recall and precison (S-RP),
{\em ii)\/} harmonic mean of recall and precision (H-RP),
{\em iii) \/} sum of density and coverage (S-DC),
{\em iv)\/}	harmonic mean of density and coverage (H-DC), and
{\em v)\/}	Jensen-Shannon distance (JSD) \cite{davaslioglu2021endtoend}.
In the GAN training, these measures are evaluated every 10 epochs and if the fidelity/diversity measure has improved, the model is saved. For the JSD metric, a lower value indicates a better model, whereas for the remaining four measures, a higher value means a better model. We used 500 epochs for each of the GAN and encoder training steps. For all three networks, we used the Adam optimizer with a learning rate of 0.0002 \cite{KB14}.

As a comparison, we evaluated the performance of a convolutional autoencoder (CAE) system which is again trained on the only inlier modulation. The CAE architecture is presented in Table~\ref{tab:cae} in the Appendix. We measured the mean and standard deviation of the reconstruction loss in the training dataset. For any test data that has a reconstruction loss larger than a fixed threshold (i.e., the mean reconstruction loss plus one standard deviation), the sample is labeled as an outlier and for those samples with a reconstruction loss smaller or equal to threshold, the sample is labeled as an inlier. Note that to have a fair assessment between performance measures and minimize the effects of network architecture, we have used a CAE model that is very similar to the AE model of the f-AnoGAN.

Table~\ref{tab:rf_anomaly_fanogan} presents the AUROC values for the anomaly detection problem employing different modulations. We observe that the performance of the anomaly detection model depends significantly on the trained (inlier) modulation in both GAN-based and CAE-based approaches. There are several reasons for this. First, unlike image datasets, this dataset (and many other RF datasets) include the effects of channel, noise, and hardware impairments, which make the classification problem fundamentally challenging. Second, some modulations in this dataset are subsets of each other in the constellation plot. For example, AM-SSB modulation is a subset of the AM-DSB modulation. In comparing the results in Table~\ref{tab:rf_anomaly_fanogan}, we see that the H-RP, JSD, and S-RP measures perform well across different modulations, whereas the S-DC and H-DC metrics did not provide good performance. We observe that anomalies can be precisely detected (achieving more than a 0.90~AUROC score) when f-AnoGAN models are trained on modulations such as the AM-DSB, CPFSK, GFSK, and WBFM using any of the S-RP, H-RP, and JSD measures. In these modulations, the f-AnoGAN method outperforms the CAE method. However, modulations such as AM-SSB and QAM16 present challenging problems as they are very similar to other classes (AM-DSB and QAM64, respectively). For these modulations, the f-AnoGAN method fails to reliably distinguish the anomalies from inliers, whereas the CAE model performs only as good as a random classifier.

\ifCLASSOPTIONonecolumn
\begin{table}
\small
\centering
\begin{tabular}{c|ccccc|c}
%        \toprule
\hline
        Modulation & S-RP & H-RP & S-DC & H-DC & JSD & CAE \\
%        \midrule
\hline
        AM-DSB & 0.963 & 0.964 & 0.962 & 0.965 & \textbf{0.968} & 0.899  \\
        AM-SSB & 0.088 &	0.088 &	0.088 &	0.088 &	\textbf{0.273} & \textbf{0.502} \\
        BPSK &	\textbf{0.797} & 0.777 &	0.349 &	0.716 &	0.439 & 0.529 \\
        8PSK &	0.558 &	0.597 &	0.540 &	0.540 &	\textbf{0.636} & 0.528 \\
        CPFSK &	0.968 &	\textbf{0.970} &	0.853 &	0.732 &	0.939 & 0.533 \\
        GFSK &	\textbf{0.983} &	0.980 &	0.821 &	0.816 &	0.917 & 0.561 \\
        PAM4 &	0.692 &	0.728 &	0.041 &	0.707 &	\textbf{0.731} & 0.587 \\
        QAM16 &	0.419 &	0.424 &	0.300 &	0.300 &	\textbf{0.447} & \textbf{0.600} \\
        QAM64 &	0.549 &	\textbf{0.624} & 0.285 &	0.285 &	0.545 & 0.597 \\
        QPSK  &	0.664 &	\textbf{0.678} &	0.547 &	0.547 &	0.559 & 0.530 \\
        WBFM  &	0.937 &	0.938 &	0.934 &	0.933 &	\textbf{0.939} & 0.824\\
        \hline
        Average &	0.693 &	\textbf{0.706} &	0.520 &	0.603 &	0.672 & 0.608 \\
 %        \bottomrule
 \hline
\end{tabular}
\caption{AUROC scores of the anomaly scores obtained using five different fidelity measures for different modulations. S-RP stands for sum of precision and recall, H-RP is the harmonic mean of precision and recall, S-DC stands for sum of density and coverage, H-DC is the harmonic mean of density and coverage, and JSD stands for Jensen-Shannon distance.}
    \label{tab:rf_anomaly_fanogan}
\end{table}
\else
\begin{table}
\vspace{2.5mm}
\small
\centering
\resizebox{0.48\textwidth}{!}{
\begin{tabular}{c|ccccc|c}
%        \toprule
\hline
        Mod. & S-RP & H-RP & S-DC & H-DC & JSD & CAE \\
%        \midrule
\hline
        AM-DSB & 0.963 & 0.964 & 0.962 & 0.965 & \textbf{0.968} & 0.899  \\
        AM-SSB & 0.088 &	0.088 &	0.088 &	0.088 &	\textbf{0.273} & \textbf{0.502} \\
        BPSK &	\textbf{0.797} & 0.777 &	0.349 &	0.716 &	0.439 & 0.529 \\
        8PSK &	0.558 &	0.597 &	0.540 &	0.540 &	\textbf{0.636} & 0.528 \\
        CPFSK &	0.968 &	\textbf{0.970} &	0.853 &	0.732 &	0.939 & 0.533 \\
        GFSK &	\textbf{0.983} &	0.980 &	0.821 &	0.816 &	0.917 & 0.561 \\
        PAM4 &	0.692 &	0.728 &	0.041 &	0.707 &	\textbf{0.731} & 0.587 \\
        QAM16 &	0.419 &	0.424 &	0.300 &	0.300 &	\textbf{0.447} & \textbf{0.600} \\
        QAM64 &	0.549 &	\textbf{0.624} & 0.285 &	0.285 &	0.545 & 0.597 \\
        QPSK  &	0.664 &	\textbf{0.678} &	0.547 &	0.547 &	0.559 & 0.530 \\
        WBFM  &	0.937 &	0.938 &	0.934 &	0.933 &	\textbf{0.939} & 0.824\\
        \hline
        Average &	0.693 &	\textbf{0.706} &	0.520 &	0.603 &	0.672 & 0.608 \\
 %        \bottomrule
 \hline
\end{tabular}
}
\caption{AUROC scores of the anomaly scores obtained using five different fidelity measures for different modulations. S-RP stands for sum of precision and recall, H-RP is the harmonic mean of precision and recall, S-DC stands for sum of density and coverage, H-DC is the harmonic mean of density and coverage, and JSD stands for Jensen-Shannon distance.}
    \label{tab:rf_anomaly_fanogan}
\end{table}
\fi

Next, we discuss some implementation details. For the WGAN-GP training, there are two important parameters that we have tuned. The first parameter is the number of discriminator (critic) iterations per generator iteration. We have tested 1, 3, and 5 iterations and found that updating the generator every 3 discriminator updates provided the best results. The second parameter is the gradient penalty coefficient $\lambda$. This parameter had a profound effect on the performance. We tested different $\lambda$ values ($\lambda=1,5,10$) and $\lambda=10$ provided the best results, which emphasizes the importance of the gradient penalty. For the f-AnoGAN model, the scaling factor $\kappa$ is a parameter that can be tuned. We have tested $\kappa = 1, 5, 10$ values and found that $\kappa = 1$ provided the best results. Finally, as the f-AnoGAN approach requires only one forward pass and does not need any iteration to infer a sample, we can obtain the average inference time on hardware, which is an important measure for the embedded implementation. Note that this is in contrast to iterative algorithms such as AnoGAN that require setting parameters such as the maximum number of iterations to obtain precise timing measures. Towards this goal, we measured the average end-to-end inference time on hardware, which includes the forward pass of the input sample to the encoder, generator, and discriminator networks, and calculating the respective losses to obtain the anomaly score. The results are evaluated on an NVIDIA GeForce RTX 3060 GPU system. The average inference time is measured as 0.005784 seconds which enables us to process 172.9 samples per second without any parallel processing.

In concluding this section, we would like to emphasize that our simulations can only be considered as very initial results. To be able to come to more in-depth conclusions, substantially more work by the research community should be carried out.

\section{Future Research Directions for GANs in NextG Communication Systems}\label{sec:GANs4Wireless}
This section is on the use of GANs for wireless applications in general terms. This is important because the original development of GANs was in the area of images while their straightforward
application to wireless can bring up questions. On the other
hand, this feature actually makes the proposed research nontrivial and
interesting. It should be understood that what we are
proposing is not a simple translation or porting of what exists in
another field to wireless communications but rather a very careful
study of how to be inspired from a development in one field
and do a careful analysis to adapt it to a completely different area. For
example, performance measures used to compare two images do not
directly translate into wireless applications. In particular,
performance measures used in image processing try to imitate human
perception. Clearly, it is questionable that they would be useful in
the kind of tasks discussed above in sections Sec~\ref{sec:sharing}--Sec~\ref{sec:security}. If this is the case, then what kind of performance measures should be employed? Since ML is
not a model-based approach, finding the answer to this question will
require experimentation and thus is not a straightforward
task. Another consideration comes from the fact that GANs are known
to be difficult to train, especially for datasets other than images. Then, is there a way this difficulty can be circumvented?
In terms of datasets, there is a perception that datasets that can be
employed for wireless applications are limited in quantity. Furthermore, in general, training of neural networks require very long training sequences. On the other hand, the datasets available are limited in size. Some researchers employ GANs to increase the size of the their training sequences in a synthetic manner. But then there is a question as to whether the generated or augmented synthetic signals contain any more information than the training set since the GAN is able to learn the statistics of the training sequence but it is not clear if this augmentation process can add any new statistics.
These are some of the questions that employing algorithms from the basic exploration of GANs to their use in wireless applications will face.

A number of %concerns
open problems remain in this domain. While a large number of problems in spectrum sharing, anomaly detection, and security exist, studies using GANs to address them have so far targeted a relatively small number of specific problems. Scaling these solutions to a larger number of problems will require significant effort. First, specific problems need to be identified, for example, in anomaly detection and security, to be addressed by ML algorithms. Second, these specific problems need to be simulated and ML algorithms need to be trained on them. Third, the computational complexity of this effort should be taken under consideration. There are many successful uses of GANs, especially in image processing. While a number of these algorithms have been successfully used in wireless applications, a number of them are still candidates for use in that setting. For example, we have discussed the use of BiGANs and f-AnoGANs for anomaly detection, however, a number of other anomaly detection techniques such as EGBAD and GANomaly have not yet been studied in the context of wireless anomaly detection. Fourth, reliable diversity and fidelity measures for GANs to quantify the quality of the synthetic data for RF applications need to be further explored. As it is difficult to conduct statistical analyses over complex and high-dimensional data, these metrics would provide the necessary reliable methods to perform hyper-parameter optimization in training GANs and fairly compare different GAN architectures. Towards this goal, we have investigated five of such metrics in this paper and showed that HR-P provided the best results for the outlier detection problem.

Yet on another subject we touched upon previously, the topic of augmentation is controversial. Some researchers strongly hold the belief that GAN-generated signals cannot be gainfully employed to augment the training dataset. However, an interesting piece of news from NVIDIA, discussed in \cite{NVIDIA}, together with a short video worthwhile seeing, states that it is possible to reduce the training sequence length 10x-20x by using data augmentation and GANs. The particular application is the creation of synthetic images from real ones and therefore the visual effect is strong. This was achieved by employing a special technique with GANs, explained in \cite{KAHLLA20}. Additional works show that, by using special techniques, it is possible to achieve data augmentation outside the statistics of the short training sequence, see, e.g., \cite{ASE18,SK19,ZLLZH20}. There exist publications that make this statement specifically for wireless applications, see, e.g., \cite{TTZL18,LLLZ18,CHHM21}.

An inevitable question arises when GANs are considered as a member of a large class of ML algorithms. As shown in this paper, they have advantages in terms performance although it is not clear that they are always the best. Considerations such as complexity, accuracy, robustness, etc. should be taken into account in comparing them with other ML algorithms for use in the wireless applications considered in this paper. To the best of our knowledge, such a study does not exist. Any step taken towards this understanding will be very valuable.

GANs can be applied to a number of emerging areas in communications. An example of this is communications in the mmWave and THz bands. As long as the computational complexity of the solution can be managed, what was discussed in this paper is applicable to communication networks in those bands. We will discuss a number of works from the literature related to this fact. Reference \cite{BA21} discusses channel estimation for very large bandwidths operating in mmWave and THz bands at low SNR. It proposes a GAN-based estimator that first learns to produce channel samples from the unknown channel distribution via training the generative network, and then uses this trained network as a prior to estimate the current channel. This estimator has been shown to work at an SNR of -5 dB better than Least Squares (LS) and Linear Minimum Mean Square Estimate (LMMSE) estimators at SNR values of 20 dB and 2.5 dB, respectively. Furthermore, this estimator reduces the required number of pilots and does not require retraining even if the number of clusters and rays change considerably. Channel estimation in the mmWave band is further discussed for the unmanned aerial vehicle (UAV) applications in \cite{ZFS21,ZFSB21} and for vehicular systems in \cite{LAT18}.

Another emerging area in communications is mobile edge computing. This is an area GANs have applications in security. As discussed in \cite{LLHJLYNM20}, GANs can be used to design a security attack for mobile edge computing. In this attack, a malicious participant can infer sensitive information from a victim participant even with just a part of shared parameters from the victim \cite{HAP17}, \cite[Fig. 13]{LLHJLYNM20}. For other applications of GANs in mobile edge computing, in security and privacy, see, e.g., \cite{HAP17,CLCYK20}.
\section{Conclusion}\label{sec:conclusion}
The proliferation of wireless services, in the forms of wireless local area networks, cellular wireless, applications of the IoT, the tactile internet, autonomous cars, drones, factory automation, and the general expectations under the umbrella of the Fourth Industrial Revolution place enormous demands on wireless spectrum. Although new spectrum in the mmWave band is already being regulated around the world, the expectations are such that spectrum scarcity will be with us beginning in the very near future. Many experts in the field expect that forms of spectrum sharing will be needed towards that purpose. It is well known that a common technique under consideration for dynamic spectrum sharing is cognitive radio which can be used to address the problem of spectrum sharing. However, conventional forms of cognitive radio will likely not be sufficient because of the enormous dynamics of the problem. It is expected that the complexity of the problem will prohibit conventional cognitive radio and forms of ML will have to be employed towards solving this very difficult problem.

This paper addressed a particular ML solution, GANs, towards that end. GANs are capable of generating ``fake'' data that cannot be differentiated from ``real'' data, and thus to some extent they capture the distribution of real signals. They are known to address competitive resource allocation problems. Furthermore, they are known to be effective in detecting and mitigating anomalous behavior. A survey of the use of GANs in spectrum sharing, anomaly detection, and addressing security concerns is provided in this paper. In several cases discussed, it is demonstrated that GANs have better performance in addressing the concerns stated above than the other ML algorithms.

It is generally accepted that wireless applications are increasing in quantity while wireless spectrum is scarce and that ML solutions to address this problem are needed. GANs provide a framework for such a set of solutions with proven success. Nevertheless, more work is needed to identify specific problems and develop successful ML solutions, including those that involve GANs.

\section{Acknowledgments}\label{sec:acknowledgments}
The authors would like to thank the anonymous reviewers whose comments improved the presentation in the paper. 
\ifCLASSOPTIONonecolumn
\clearpage
\newpage
\fi
\section*{Appendix}\label{appendix}
In this section, we provide the parameters of the f-AnoGAN and CAE used in the simulations in Sec.~\ref{sec:initres}.

\ifCLASSOPTIONonecolumn
\begin{table}[th!]
    \centering
    \small
    \begin{tabular}{l|l}
    \toprule
    Layer Name &  Details  \\ \midrule
    Reshape layer & $(N,100)$ to $(N,100,1,1)$ \\
    2D transposed convolution layer & 1024 filters, kernel size of (2,4), stride of (1,1), no padding \\
    Batch normalization &  \\
    Leaky-ReLU & $\alpha=0.2$ \\
    2D transposed convolution layer & 512 filters, kernel size of (2,4), stride of (2,2), padding \\
    Batch normalization &  \\
    Leaky-ReLU & $\alpha=0.2$ \\
    2D transposed convolution layer & 256 filters, kernel size of (2,4), stride of (2,2), padding \\
    Batch normalization &  \\
    Leaky-ReLU & $\alpha=0.2$ \\
    2D transposed convolution layer & 128 filters, kernel size of (2,4), stride of (2,2), padding \\
    Batch normalization &  \\
    Leaky-ReLU & $\alpha=0.2$ \\
    2D transposed convolution layer & 64 filters, kernel size of (2,4), stride of (2,2), padding \\
    Batch normalization &  \\
    Leaky-ReLU & $\alpha=0.2$ \\
    2D transposed convolution layer & 1 filter; kernel size of (2,4), stride of (2,2), padding \\
    \bottomrule
    \end{tabular}
    \caption{Generator architecture.}
    \label{tab:generator_model}
\end{table}
\else
\begin{table}[ht]
    \centering
    \small
    \begin{tabular}{l|l}
    \toprule
    Layer Name &  Details  \\ \midrule
    Reshape layer & $(N,100)$ to $(N,100,1,1)$ \\
    2D transposed  & 1024 filters, kernel size of  \\ convolution layer &(2,4), stride of (1,1), no padding \\
    Batch normalization &  \\
    Leaky-ReLU & $\alpha=0.2$ \\
    2D transposed   & 512 filters, kernel size of \\ convolution layer & (2,4), stride of (2,2), padding \\
    Batch normalization &  \\
    Leaky-ReLU & $\alpha=0.2$ \\
    2D transposed  & 256 filters, kernel size of \\convolution layer & (2,4), stride of (2,2), padding \\
    Batch normalization &  \\
    Leaky-ReLU & $\alpha=0.2$ \\
    2D transposed   & 128 filters, kernel size of\\ convolution layer & (2,4), stride of (2,2), padding \\
    Batch normalization &  \\
    Leaky-ReLU & $\alpha=0.2$ \\
    2D transposed   & 64 filters, kernel size of\\ convolution layer& (2,4), stride of (2,2), padding \\
    Batch normalization &  \\
    Leaky-ReLU & $\alpha=0.2$ \\
    2D transposed   & 1 filter; kernel size of\\ convolution layer& (2,4),stride of (2,2), padding \\
    \bottomrule
    \end{tabular}
    \caption{Generator architecture.}
    \label{tab:generator_model}
\end{table}
\fi

\ifCLASSOPTIONonecolumn
\begin{table}[t!]
    \centering
    \small
    \begin{tabular}{l|l}
    \toprule
    Layer Name &  Details  \\ \midrule
    2D convolution layer &  64 filters, kernel size of (2,4), stride of (2,2), padding  \\
    Leaky-ReLU & $\alpha=0.2$ \\
    2D convolution layer &  128 filters, kernel size of (2,4), stride of (2,2), padding  \\
    Batch normalization & \\
    Leaky-ReLU & $\alpha=0.2$ \\
    2D convolution layer &  256 filters, kernel size of (2,4), stride of (2,2), padding  \\
    Batch normalization & \\
    Leaky-ReLU & $\alpha=0.2$ \\
    2D convolution layer &  512 filters, kernel size of (2,4), stride of (2,2), padding  \\
    Batch normalization & \\
    Leaky-ReLU & $\alpha=0.2$ \\
    2D convolution layer &  1024 filters, kernel size of (2,4), stride of (2,2), padding  \\
    Reshape layer & $(N,1024,2,4)$ to $(N,8192)$ \\
    Linear layer & $8192$ neurons to $1$ neuron \\
    \bottomrule
    \end{tabular}
    \caption{Discriminator architecture.}
    \label{tab:discriminator_model}
\end{table}
\else
\begin{table}[ht]
    \centering
    \small
    \begin{tabular}{l|l}
    \toprule
    Layer Name &  Details  \\ \midrule
    2D convolution layer &  64 filters, kernel size of (2,4), \\& stride of (2,2), padding  \\
    Leaky-ReLU & $\alpha=0.2$ \\
    2D convolution layer &  128 filters, kernel size of (2,4),\\& stride of (2,2), padding  \\
    Batch normalization & \\
    Leaky-ReLU & $\alpha=0.2$ \\
    2D convolution layer &  256 filters, kernel size of (2,4),\\& stride of (2,2), padding  \\
    Batch normalization & \\
    Leaky-ReLU & $\alpha=0.2$ \\
    2D convolution layer &  512 filters, kernel size of (2,4),\\& stride of (2,2), padding  \\
    Batch normalization & \\
    Leaky-ReLU & $\alpha=0.2$ \\
    2D convolution layer &  1024 filters, kernel size of (2,4),\\& stride of (2,2), padding  \\
    Reshape layer & $(N,1024,2,4)$ to $(N,8192)$ \\
    Linear layer & $8192$ neurons to $1$ neuron \\
    \bottomrule
    \end{tabular}
    \caption{Discriminator architecture.}
    \label{tab:discriminator_model}
\end{table}
\fi 

\ifCLASSOPTIONonecolumn
\begin{table}[t!]
    \centering
    \small 
    \begin{tabular}{c|c}
    \toprule
    Layer Name &  Details  \\ \midrule
    2D convolution layer &  64 filters, kernel size of (2,4), stride of (2,2), padding  \\
    Leaky-ReLU & $\alpha=0.2$ \\
    2D convolution layer &  64 filters, kernel size of (2,4), stride of (2,2), padding  \\
    Batch normalization & \\
    Leaky-ReLU & $\alpha=0.2$ \\
    2D convolution layer &  128 filters, kernel size of (2,4), stride of (2,2), padding  \\
    Batch normalization & \\
    Leaky-ReLU & $\alpha=0.2$ \\
    2D convolution layer &  256 filters, kernel size of (2,4), stride of (2,2), padding  \\
    Batch normalization & \\
    Leaky-ReLU & $\alpha=0.2$ \\
    2D convolution layer &  512 filters, kernel size of (2,4), stride of (2,2), padding  \\
    Batch normalization & \\
    Leaky-ReLU & $\alpha=0.2$ \\
    2D convolution layer &  1024 filters, kernel size of (2,4), stride of (2,2), padding  \\
    Reshape layer & $(N,1024,2,2)$ to $(N,100)$ \\
    Linear layer & 4096 neurons to $1$ neuron \\
    Tanh &  \\
    \bottomrule
    \end{tabular}
    \caption{Encoder architecture of the f-AnoGAN system.}
    \label{tab:encoder_model}
\end{table}
\else
\begin{table}[ht!]
    \centering
    \small 
    \begin{tabular}{c|c}
    \toprule
    Layer Name &  Details  \\ \midrule
    2D convolution layer &  64 filters, kernel size of (2,4),\\& stride of (2,2), padding  \\
    Leaky-ReLU & $\alpha=0.2$ \\
    2D convolution layer &  64 filters, kernel size of (2,4),\\& stride of (2,2), padding  \\
    Batch normalization & \\
    Leaky-ReLU & $\alpha=0.2$ \\
    2D convolution layer &  128 filters, kernel size of (2,4),\\& stride of (2,2), padding  \\
    Batch normalization & \\
    Leaky-ReLU & $\alpha=0.2$ \\
    2D convolution layer &  256 filters, kernel size of (2,4),\\& stride of (2,2), padding  \\
    Batch normalization & \\
    Leaky-ReLU & $\alpha=0.2$ \\
    2D convolution layer &  512 filters, kernel size of (2,4),\\& stride of (2,2), padding  \\
    Batch normalization & \\
    Leaky-ReLU & $\alpha=0.2$ \\
    2D convolution layer &  1024 filters, kernel size of (2,4),\\& stride of (2,2), padding  \\
    Reshape layer & $(N,1024,2,2)$ to $(N,100)$ \\
    Linear layer & 4096 neurons to $1$ neuron \\
    Tanh &  \\
    \bottomrule
    \end{tabular}
    \caption{Encoder architecture of the f-AnoGAN system.}
    \label{tab:encoder_model}
\end{table}
\fi 

\ifCLASSOPTIONonecolumn
\begin{table}[t!]
    \centering
    \small 
    \begin{tabular}{c|c}
    \toprule
    Layer Name &  Details  \\ \midrule
    2D convolution layer &  64 filters, kernel size of (2,4), stride of (2,2), padding  \\
    Leaky-ReLU & $\alpha=0.2$ \\
    2D convolution layer &  64 filters, kernel size of (2,4), stride of (2,2), padding  \\
    Batch normalization & \\
    Leaky-ReLU & $\alpha=0.2$ \\
    2D convolution layer &  128 filters, kernel size of (2,4), stride of (2,2), padding  \\
    Batch normalization & \\
    Leaky-ReLU & $\alpha=0.2$ \\
    2D convolution layer &  256 filters, kernel size of (2,4), stride of (2,2), padding  \\
    Batch normalization & \\
    Leaky-ReLU & $\alpha=0.2$ \\
    2D convolution layer &  512 filters, kernel size of (2,4), stride of (2,2), padding  \\
    Batch normalization & \\
    Leaky-ReLU & $\alpha=0.2$ \\
    2D convolution layer &  1024 filters, kernel size of (2,4), stride of (2,2), padding  \\
    Reshape layer & $(N,1024,2,2)$ to $(N,100)$ \\ \hline
    Reshape layer & $(N,100)$ to $(N,1024,2,2)$ \\ 
    2D transposed convolution layer & 512 filters, kernel size of (1,2), stride of (1,2), padding \\
    Leaky-ReLU & $\alpha=0.2$ \\
    Batch normalization & \\
    2D transposed convolution layer & 256 filters, kernel size of (1,2), stride of (1,2), no padding \\
    Leaky-ReLU & $\alpha=0.2$ \\
    Batch normalization & \\
    
    2D transposed convolution layer & 128 filters, kernel size of (1,2), stride of (1,2), no padding \\
    Leaky-ReLU & $\alpha=0.2$ \\
    Batch normalization & \\
    
    2D transposed convolution layer & 64 filters, kernel size of (1,2), stride of (1,2), no padding \\
    Leaky-ReLU & $\alpha=0.2$ \\
    Batch normalization & \\
    
    2D transposed convolution layer & 64 filters, kernel size of (1,2), stride of (1,2), no padding \\
    Leaky-ReLU & $\alpha=0.2$ \\
    2D transposed convolution layer & 1 filter, kernel size of (1,2), stride of (1,2), no padding \\
    
    \bottomrule
    \end{tabular}
    \caption{Convolutional Autoencoder (CAE) architecture.}
    \label{tab:cae}
\end{table}
\else
\begin{table}[ht!]
    \centering
    \small 
    \begin{tabular}{c|c}
    \toprule
    Layer Name &  Details  \\ \midrule
    2D convolution layer &  64 filters, kernel size of (2,4), \\&stride of (2,2), padding  \\
    Leaky-ReLU & $\alpha=0.2$ \\
    2D convolution layer &  64 filters, kernel size of (2,4), \\&stride of (2,2), padding  \\
    Batch normalization & \\
    Leaky-ReLU & $\alpha=0.2$ \\
    2D convolution layer &  128 filters, kernel size of (2,4),\\& stride of (2,2), padding  \\
    Batch normalization & \\
    Leaky-ReLU & $\alpha=0.2$ \\
    2D convolution layer &  256 filters, kernel size of (2,4),\\& stride of (2,2), padding  \\
    Batch normalization & \\
    Leaky-ReLU & $\alpha=0.2$ \\
    2D convolution layer &  512 filters, kernel size of (2,4),\\& stride of (2,2), padding  \\
    Batch normalization & \\
    Leaky-ReLU & $\alpha=0.2$ \\
    2D convolution layer &  1024 filters, kernel size of (2,4),\\& stride of (2,2), padding  \\
    Reshape layer & $(N,1024,2,2)$ to $(N,100)$ \\ \hline
    Reshape layer & $(N,100)$ to $(N,1024,2,2)$ \\ 
    2D transposed  & 512  filters, kernel size of (1,2),\\ convolution layer & stride of (1,2), padding \\
    Leaky-ReLU & $\alpha=0.2$ \\
    Batch normalization & \\
    2D transposed  & 256 filters, kernel size of (1,2),\\convolution layer & stride of (1,2), no padding \\
    Leaky-ReLU & $\alpha=0.2$ \\
    Batch normalization & \\
    
    2D transposed  & 128 filters, kernel size of (1,2),\\ convolution layer & stride of (1,2), no padding \\
    Leaky-ReLU & $\alpha=0.2$ \\
    Batch normalization & \\
    
    2D transposed  & 64 filters, kernel size of (1,2),\\ convolution layer& stride of (1,2), no padding \\
    Leaky-ReLU & $\alpha=0.2$ \\
    Batch normalization & \\
    
    2D transposed  & 64 filters, kernel size of (1,2),\\ convolution layer& stride of (1,2), no padding \\
    Leaky-ReLU & $\alpha=0.2$ \\
    2D transposed  & 1 filter, kernel size of (1,2),\\ convolution layer & stride of (1,2), no padding \\
    \bottomrule
    \end{tabular}
    \caption{Convolutional Autoencoder (CAE) architecture.}
    \label{tab:cae}
\end{table}
\fi 

\ifCLASSOPTIONonecolumn
\clearpage
\newpage
\fi
\small
\bibliographystyle{IEEEtran}
\bibliography{References/References}
\end{document}